\documentclass{article}

\PassOptionsToPackage{numbers, compress}{natbib}
\usepackage[rightcaption]{sidecap}

    \usepackage[preprint]{neurips_2025}
\usepackage[utf8]{inputenc} % allow utf-8 input
\usepackage[T1]{fontenc}    % use 8-bit T1 fonts
\usepackage{url}            % simple URL typesetting
\usepackage{booktabs}       % professional-quality tables
\usepackage{amsfonts}       % blackboard math symbols
\usepackage{nicefrac}       % compact symbols for 1/2, etc.
\usepackage{microtype}      % microtypography
\usepackage[table]{xcolor}
\usepackage{makecell}
\usepackage{wrapfig}
\usepackage{graphicx}
\usepackage{subcaption}
\usepackage{booktabs}
\usepackage{amsmath}
\usepackage{amssymb}
\usepackage{float}
\usepackage{enumitem}
\usepackage{multirow}
\usepackage[mathscr]{eucal}
\usepackage[export]{adjustbox}
\usepackage{multicol,lipsum}
\usepackage{arydshln}
\usepackage{pifont}
\usepackage{booktabs}
\usepackage{bbding}
\usepackage{multirow}
\usepackage{colortbl}
\usepackage{graphicx}
\usepackage{float}
\usepackage[ruled,vlined]{algorithm2e}
\usepackage{pythonhighlight} 
\usepackage{fontawesome5}
\usepackage[accsupp]{axessibility}  % Improves PDF readability for those with disabilities.

\usepackage{hyperref}

\def\0{{\bf 0}}
\def\1{{\bf 1}}

\definecolor{purple}{rgb}{0.56,0.27,0.68}
\definecolor{red}{rgb}{0.95,0.4,0.4}
\definecolor{purered}{rgb}{1,0,0}
\definecolor{blue}{rgb}{0.4,0.4,0.95}
\definecolor{darkblue}{rgb}{0,0,0.8}
\definecolor{lightblue}{rgb}{127,153,240}
\definecolor{grey}{rgb}{0.6,0.6,0.6}
\definecolor{col1}{RGB}{232, 161, 148}
\definecolor{col11}{RGB}{255, 228, 228}
\definecolor{col2}{RGB}{148, 187, 232}
\definecolor{col33}{RGB}{206, 239, 255}
\definecolor{col3}{RGB}{233, 255, 245}
\definecolor{lightgrey}{rgb}{0.85,0.85,0.85}
\definecolor{lightyellow}{RGB}{255,195,78}
\definecolor{lightlightgrey}{rgb}{0.9,0.9,0.9}
\definecolor{verylightBG}{rgb}{0.9,0.99,0.99}
\definecolor{darkgreen}{rgb}{0., 0.85, 0.5}
\definecolor{tfs_lp}{RGB}{130, 176, 210}
\definecolor{tfs_pt}{RGB}{232, 197, 31}
\definecolor{tfs_ft}{RGB}{142, 207, 201}
\definecolor{tfs_ct}{RGB}{250,95,111}

\definecolor{fig2_random}{HTML}{038355}
\definecolor{fig2_entropy}{HTML}{6b9ac8}
\definecolor{fig2_coreset}{HTML}{f8ac8c}
\definecolor{fig2_badge}{HTML}{accf78}
\definecolor{fig2_pcb}{HTML}{c497b2}
\definecolor{fig2_alfa}{HTML}{3d5e80}
\definecolor{fig2_logo}{HTML}{7a8c7b}
\definecolor{fig3_tfs}{HTML}{fa7f6f}

\definecolor{gtred}{RGB}{204, 0, 0}
\definecolor{predgreen}{RGB}{31, 237, 31}
\definecolor{figGreen}{RGB}{56, 118, 29}

\graphicspath{{./figures/}}   %  image

\lstnewenvironment{pythonic}[1][]{\lstset{style=mypython, frame=none, #1}}{}

\usepackage[capitalize]{cleveref}
\crefname{section}{Sec.}{Secs.}
\Crefname{section}{Section}{Sections}
\Crefname{table}{Table}{Tables}
\crefname{table}{Tab.}{Tabs.}

\title{Active Learning via Vision-Language Model Adaptation
with Open Data}

\author{%
Tong Wang$^{1}$, Jiaqi Wang$^{2}$,
Shu Kong$^{1,3,}\thanks{Corresponding author.}$ \\
$^1$University of Macau, 
$^2$Shanghai AI Lab,
$^3$Institute of Collaborative Innovation 
  \\
  \href{https://leowangtong.github.io/ALOR}{project webpage}
  \vspace{-5mm}
}

\begin{document}
\maketitle

\begin{abstract}
Pretrained on web-scale open data, Vision-Language Models (VLMs) offer powerful capabilities for solving downstream tasks, after being adapted to task-specific labeled data.
However, data labeling can be expensive, especially when domain expertise is required. Active Learning (AL) aims to reduce this expense by strategically selecting the most informative data for labeling and model training. Recent AL methods have explored open-source VLMs but have yet to fully leverage publicly available open data, such as datasets used for pretraining VLMs.
In this work, we propose leveraging VLM's pretraining data by retrieving samples closely related to the downstream task, using them to augment the task-specific data for AL. As expected, incorporating this data into existing AL methods leads to significant performance improvements.
Given that our method exploits open-source VLM and open data, we refer to it as \emph{Active Learning with Open Resources (ALOR)}.
Additionally, most VLM-based AL methods use prompt tuning (PT) for model adaptation, likely due to its ability to directly utilize pretrained parameters and the assumption that doing so reduces the risk of overfitting to limited labeled data. We rigorously compare popular adaptation approaches, including linear probing (LP), finetuning (FT), and contrastive tuning (CT). Our results reveal two key findings: (1) all adaptation approaches benefit significantly from incorporating retrieved data,
and (2) CT consistently outperforms other adaptation approaches  regardless what AL methods to apply, even when not using retrieved data.
Further analysis of retrieved data reveals a naturally imbalanced distribution of task-relevant classes, exposing inherent biases within the VLM. This insight motivates our novel \emph{Tail First Sampling (TFS)} strategy for AL, an embarrassingly simple yet effective method that prioritizes sampling data from underrepresented classes to label.
Extensive experiments on standard benchmark datasets demonstrate that our final method, contrastively finetuning VLM on both retrieved and TFS-selected labeled data, achieves state-of-the-art performance, significantly surpassing existing methods.

\end{abstract}

{
\begin{SCfigure}[][t]
  \centering
  \includegraphics[width = 0.5\textwidth, clip=true,trim = 0mm 0mm 0mm 0mm]{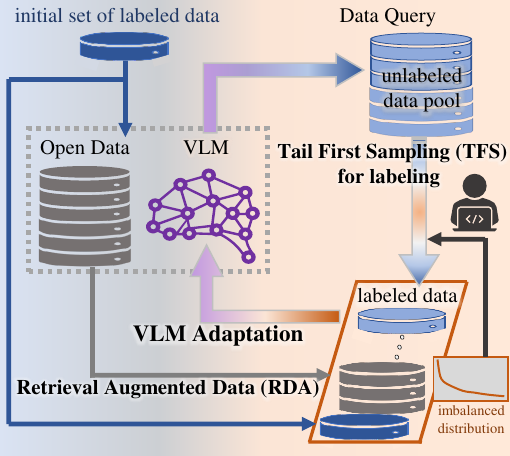}
  \hfill
  \vspace{5mm}
  \caption{\small
    {\bf Active Learning with Open Resources (ALOR)} exploits open-source VLM and open data (especially VLM's pretraining data), unlike recent AL methods which use only the former.
    Specifically, for task-specific class names, ALOR retrieves relevant pretraining data to augment the limited number of task-specific labeled data.
    The retrieved data not only reveals an imbalanced distribution but also unveils how the VLM is biased and how task-specific unlabeled data is (similarly) imbalanced.
    Leveraging these insights, ALOR adopts \emph{Tail First Sampling (TFS)} that prioritizes sampling unlabeled examples data from underrepresented classes to label.
    ALOR adapts the VLM by finetuning it over both the retrieved and TFS-selected labeled data, achieving significantly better performance than prior arts by $+$7\% accuracy on multiple benchmarks (Table~\ref{tab:benchmarking-results}).
} 
\label{fig:framework}
\vspace{-4mm}
\end{SCfigure}
}

\section{Introduction}
\label{sec:intro}

Pretrained Vision-Language Models (VLMs)~\cite{radford2021learning, wei2021aligning} demonstrate strong zero-shot performance on various visual recognition~\cite{parashar2024neglected}.
Yet, to excel in a downstream task, a VLM requires adaptation, e.g., through finetuning on labeled task-specific data. However, data labeling is expensive and often demands domain expertise. 
Pool-based Active Learning (AL) aims to reduce this expense by strategically selecting  informative unlabeled data for labeling and model training~\cite{Ash2020Badge,ash2021gone,kim2023re,alfamix,sener2018coreset}.

{\bf Status Quo.}
The core to AL is data selection strategies, which are commonly based on  uncertainty~\cite{gal2017deep, holub2008entropy, kirsch2019batchbald}, 
diversity~\cite{alfamix, sener2018coreset},
or both~\cite{hsu2015active, huang2010active, Ash2020Badge}. 
Recent methods have exploited open-source VLMs for data selection and adapted them for AL~\cite{safaei2024active, bang2024active}.
Yet, it primarily adopts prompt tuning (PT)~\cite{zhou2022coop, khattak2023maple, zhou2022cocoop} as an VLM adaptation approach,
likely due to its ability to directly leverage the pretrained parameters to reduce the risk of overfitting to the limited labeled data.
That said, it is underexplored whether other adaptation approaches such as finetuning can perform better than PT.
Furthermore, despite exploiting an open-source VLM, contemporary AL methods have yet to leverage publicly-available open data, e.g., the  VLM's pretraining data~\cite{Liu_2023_react, liu2025few} or Internet data~\cite{pmlr-v202-InternetExplorer}, but related domains have done so such as few-shot learning.

{\bf Technical Insights.}
For AL, in addition to using VLM~\cite{safaei2024active, bang2024active},
we embrace open data, particularly the VLM's open-source pretraining data~\cite{laion400m,laion5b} (for  reproduction).
We retrieve pretraining examples relevant to the downstream task and use them to augment labeled data. Unsurprisingly, adapting the VLM using both retrieved and labeled data greatly enhances AL (\Cref{tab:table_Comparison_RDA}).
Moreover,
we rigorously compare different adaptation approaches including prompt tuning (PT), linear probing (LP), tuning (FT), and contrastive finetuning (CT).
Our results show that CT resoundingly outperforms the others, regardless whether to exploit the retrieved data and what AL methods to use.
Importantly, the retrieved data reveals imbalanced distributions of task-relevant classes, implying how the VLM is biased and how the unlabeled task-specific data is similarly imbalanced  (as they are sampled in the real world).
This insight motivates our proposed \emph{Tail First Sampling (TFS)} strategy, which is an embarrassingly simple yet effective AL method that prioritizes sampling unlabeled data for underrepresented classes to label.
We call our method \emph{Active Learning with Open Resources (ALOR)},
as it exploits both open-source VLM and open data.
We summarize ALOR in Fig.~\ref{fig:framework}, which consists of three major components: retrieval-based data augmentation (RDA), VLM adaptation, and the TFS strategy.
Our ALOR significantly outperforms prior arts (\Cref{tab:benchmarking-results}).

{\bf Contributions.}
We make three major contributions: 
\begin{enumerate}[leftmargin=15pt, topsep=0pt, itemsep=5pt,parsep=-2pt]
    \item 
    We study AL by embracing both a VLM and its pretraining data (instantiating the \emph{open data}). Particularly,
    we present \emph{retrieval-based data augmentation (RDA)} to retrieve VLM's pretraining data relevant to the downstream task, greatly enhancing existing AL methods.
    \item
    We observe that the retrieved data follows imbalanced distributions, implying how the  VLM is biased and how unlabeled task-specific data is similarly imbalanced distributed.
    This motivate our simple yet novel \emph{Tail First Sampling (TFS)} strategy that prioritizes rare classes in data selection. Extensive experiments show that TFS outperforms prior AL methods.
    \item
    We rigorously compare different VLM adaptation approaches including finetuning (FT), contrastive tuning (CT), linear probing (LP) and prompt tuning (PT).
    We show that CT significantly outperforms the others, even when retrieved data is not used.
    Our final method assembles CT, RDA and TFS, achieving the state-of-the-art on five benchmarks.
\end{enumerate}

\section{Related Work}
\label{sec:related-work}

{\bf Active Learning (AL).}
Meticulous data labeling is crucial to machine-learning solutions for real-world applications, but  data labeling is often tedious, laborious, and costly that requires domain expertise~\cite{sheng2008get, hendrycks2018using, nettleton2010study}.
AL mitigates this cost by strategically selecting the most informative unlabeled data to label during model training~\cite{settles2009active, ren2021survey}.
Recent AL methods usually rely on uncertainty~\cite{holub2008entropy}, diversity~\cite{roth2006margin,gal2017deep,wang2014new}, or both~\cite{Ash2020Badge}.
Uncertainty-based methods select unlabeled examples which have high uncertainties of model's predictions, e.g., high entropy~\cite{holub2008entropy}.
Such methods concentrate on forecasting the likelihood of classes while ignoring the distribution of the unlabeled data~\cite{roth2006margin,gal2017deep,wang2014new}. 
In contrast, diversity-based methods consider the distribution of the unlabeled data pool by selecting examples that can cover the unlabeled data distribution~\cite{sener2018coreset,yang2015multi}.
Early AL methods exploit deep models pretrained on ImageNet~\cite{Choi_2021_ICCV,Liu_2021_InfluenceSelection,Yoo_2019_LearningLoss}; recent AL methods leverage pretrained VLMs~\cite{bang2024active, safaei2024active}, particularly adopting prompt tuning with frozen visual and text encoders to select unlabeled examples.
Moving on with the AL literature, our work further embraces VLM's pretraining data in addition to the VLM itself and rigorously studies different adaptation approaches other than prompt tuning.

{\bf Vision-Language Models (VLMs)} 
consists of visual and text encoders that are jointly pretrained on web-scale  image-text paired data in a contrastive manner \cite{radford2021learning, jia2021scaling, li2022blip, li2021align}.
VLMs excel in various downstream task, such as visual captioning~\cite{chen2022visualgpt, sun2024alpha, wang2025cogvlm}, visual recognition~\cite{radford2021learning, parashar2024neglected}, visual question answering~\cite{guo2023images, sun2024alpha}, etc.
Owing to the outstanding capabilities of VLMs,
downstream tasks adapt them over labeled task-specific data~\cite{chen2022visualgpt, liu2025few}.
Various adaptation approaches exist such as prompt tuning (PT)~\cite{zhou2022coop,  khattak2023maple, roy2024consistencyguided,zhou2022cocoop}, 
linear probing (LP)~\cite{radford2021learning, lin2023multimodality},
finetuning (FT)~\cite{liu2025few, kumar2022finetuning}, and contrastive tuning (CT)~\cite{goyal2023flyp}.
To mitigate data labeling cost, recent AL methods leverage VLMs for unlabeled data selection along with VLM adaptation~\cite{bang2024active, safaei2024active}.
Yet, these methods turn to PT to adapt VLMs~\cite{bang2024active, safaei2024active}, likely because PT allows direct utilization of pretrained VLMs that reduces the risk of overfitting to limited labeled data.
Our work rigorously compares different adaptation approaches for AL and finds that CT resoundingly outperforms other approaches.

{\bf Retrieval-based Data Augmentation (RDA)} retrieves publicly-available open data relevant to a downstream task to help make predictions.
For example,
language models adopt RDA to enhance various complex reasoning tasks~\cite{pmlr-v119-guu20a,NEURIPS2020_6b493230}.
Recently, computer vision community uses RDA in not only generative modeling or inference~\cite{chen2022re, blattmann2022retrieval}
but also in model training to better solve downstream tasks such as few-shot recognition~\cite{liu2025few, pmlr-v202-InternetExplorer, NEURIPS2023_NeuralPriming, Liu_2023_react, parashar2024neglected}.
To the best of our knowledge,
RDA has been unexplored in AL.
We show that RDA significantly improves existing AL methods.
Furthermore, 
RDA reveals long-tailed distributions of real-world data belonging to the task-specific classes and and implies how pretrained VLMs are biased on these classes.
This observation motivates our novel Tail-First Sampling (TFS) strategy that prioritizes sampling unlabeled data for underrepresented classes to enhance AL.

\section{Active Learning with Open Resources}
\label{sec:ALOR}

\begin{wrapfigure}{r}{0.5\textwidth}
\centering
\vspace{-7mm}
\includegraphics[width = 0.5\textwidth, clip=true,trim = 0mm 0mm 0mm 0mm]{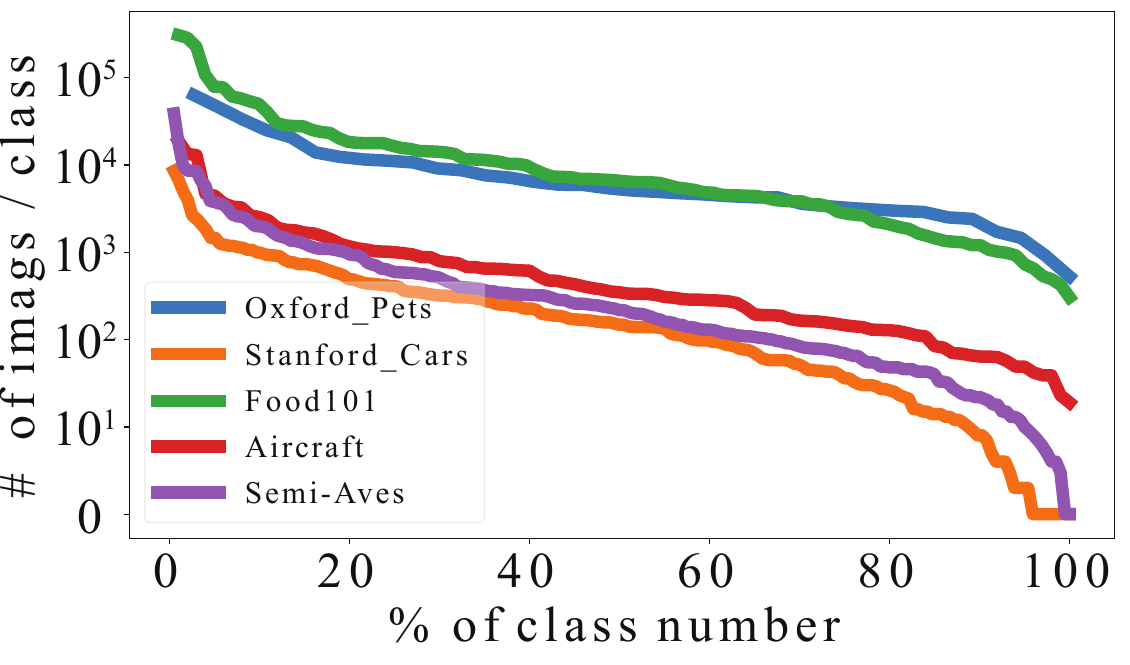}
\vspace{-6.5mm}
\caption{\small
  Retrieved data from a VLM's pretraining dataset (i.e., LAION-400M~\cite{laion400m}) follows long-tailed distributions w.r.t classes concerned by specific tasks.
  They imply how the VLM is potentially biased as being trained on such data. They also indicate how unlabeled task-specific data is similarly imbalanced distributed.
  This motivates our simple and novel \emph{Tail-First Sampling (TFS)} strategy that prioritizes sampling data for underrepresented classes. Fig.~\ref{fig:TFS_workflow} 
  depicts the workflow of TFS.}
\label{fig:long-tail-distribution}
\vspace{-9mm}
\end{wrapfigure}

We introduce our method \emph{Active Learning with Open Resources (ALOR)}, specifically embracing open-source Vision-Language Model (VLM) and open data instantiated by the VLM's pretraining dataset.
Fig.~\ref{fig:framework} depicts the conceptual framework of ALOR and \Cref{alg:tfs} presents its pseudo code.
ALOR has three important components, as presented below.

\subsection{Retrieval-based Data Augmentation}
\label{ssec:RDA}

Retrieval-based Data Augmentation (RDA) is shown to greatly enhance zero-shot recognition~\cite{Liu_2023_react, NEURIPS2023_NeuralPriming, parashar2024neglected} and few-shot recognition~\cite{liu2025few}, as it enables models to be finetuned over abundant retrieved data relevant to the downstream task.
To retrieve task-relevant data, especially from a VLM's pretraining dataset of paired images and captions,
some methods rely on feature-matching that obtain examples that have high similarities between their features and target class names~\cite{Liu_2023_react, NEURIPS2023_NeuralPriming}.
Yet, computing features of VLM's pretraining data (e.g., 400M images in LAION-400M dataset~\cite{laion400m}) is prohibitively expensive,
as it requires not only a huge compute resource but also a large bandwidth and storage to host hundreds of millions of images.
In contrast, a recent RDA approach~\cite{parashar2024neglected, liu2025few} uses string matching for retrieval, achieving significant speedup and performance gains. 
It has three steps: (1) find relevant pretraining texts via string matching for target class names, (2) fetch the corresponding images,
and (3) filter out outlier images based on feature similarities between class names and retrieved images.
This string matching-based approach is significantly more efficient than feature similarity matching-based methods.
Importantly, it achieves better performance on visual recognition as it enables using different synonyms to retrieve more diverse images for each class~\cite{liu2025few}.
As our work is not intended to propose new RDA methods, we adopt this string matching-based RDA approach~\cite{parashar2024neglected, liu2025few} to retrieve VLM's data for AL.
For a specific task and its concerned class names, we retrieve relevant data from VLM's pretraining dataset (i.e., LAION-400M used in this work).

We use the retrieved data to augment the limited amount of selectively labeled data in each round of AL. 
We show that RDA greatly enhances all the compared AL methods (\Cref{tab:table_Comparison_RDA}).
Nevertheless, the retrieved data reveals imbalanced distributions of real-world data pertaining to the classes concerned by a downstream task (\cref{fig:long-tail-distribution}).
Importantly, the imbalanced distributions imply how the pretrained VLM is biased as it is pretrained on such imbalanced data.
\cref{fig:visual_comparison_performance} shows per-class accuracy which is rather biased after adapting an VLM over the mix of retrieved data and balanced set of initially labeled data.
Next, we present a novel and simple AL method to address these issues.

% \lipsum[1-2]
{
\begin{algorithm}[t] 
\footnotesize
% \DontPrintSemicolon
\SetAlgoLined
\SetNoFillComment
\LinesNotNumbered
% \SetInd{0mm}{0mm}
% \Setvlineskip{10mm}
% \SetAlgoInsideSkip{10mm}
\caption{\small Active Learning with Open Resources (ALOR)}
\label{alg:tfs}
\textbf{Input:}    
    {Open data $\mathcal{O}$, 
    initial labeled dataset  $\mathcal{L}_{0}$, unlabeled data pool $\mathcal{U}$, total round $R$ for AL, 
    budget $N$ in each round, $Oracle$, a pretrained VLM model $M_0$.}

            {$\mathcal{L}_{0} := \mathcal{L}_{0} \cup $} retrieve$(M_0,\mathcal{L}_{0}, \mathcal{O})$ \footnotesize{\textcolor{gray}{\# retrieve open data using the initial labeled set and VLM ($\S$\ref{ssec:RDA})}} \\
            {$M_{0} :=$ adapt$(M_0, \mathcal{L}_{0}$}) \footnotesize{\textcolor{gray}{\# adapt VLM on the initial set of labeled data including retrieved examples ($\S$\ref{ssec:adaptation})}} \\
    
    \For {$r=1, 2, ..., R$} 
        {
            $\mathcal{L}_{r} := \mathcal{L}_{r-1}$ \\
            \For {$x_i \in \mathcal{U}$}
            {
                \footnotesize{$e_i = $ entropy$(M_{r-1}(x_i))$} \footnotesize{\textcolor{gray}{\# compute entropy for unlabeled data}}\\
                \footnotesize{$\hat{y}_i = \arg\max M_{r-1}(x_i)$} \footnotesize{\textcolor{gray}{\# predict labels for unlabeled data}} \\
            }            

            \footnotesize{\textcolor{gray}{\# Tail-First Sampling  ($\S$~\ref{ssec:TFS})}}\\
            \For {$j=1, ..., N$} 
                {
                {${k} ^*= \arg\min\limits_{k=1\dots K} |S_k|$} \footnotesize{\textcolor{gray}{\# find the rarest class $k$ in the labeled set;
                }
                \footnotesize{\textcolor{gray}{$S_{k}$ is the set of class-$k$ data}} \\

                \footnotesize{$i^* =\arg\max\limits_i e_i$,  \ s.t. \  $\hat y_i=k^*$
                }  
                \footnotesize{\textcolor{gray}{\# select the unlabeled datum indexed by $p$ with the largest entropy}} \\

                $(x_{i^*},y_{i^*}) = oracle((x_{i^*}))$ \footnotesize{\textcolor{gray}{\# label  by the oracle}}\\
                $\mathcal{L}_{r} = \mathcal{L}_{r} + \{(x_{i^*},y_{i^*})\}$ \footnotesize{\textcolor{gray}{\# add to labeled set}}\\
               
                $\mathcal{U} = \mathcal{U} - \{(x_{i^*},y_{i^*})\}$  \footnotesize{\textcolor{gray}{\# remove from unlabeled set}}\\
                }
            {$M_{r}$= adapt$(M_{r-1}, \mathcal{L}_{r}$}) 
            \footnotesize{\textcolor{gray}{\# VLM adaptation ($\S$\ref{ssec:adaptation})}; }
            \footnotesize{\textcolor{gray}{adapt VLM with all labeled data}} \\
            }
        }
\textbf{Output:}  $M_R$
\end{algorithm}
}

{
\begin{SCfigure}[][b]
  \centering
  \includegraphics[width = 0.5\textwidth, clip=true,trim = 0mm 0mm 0mm 0mm]{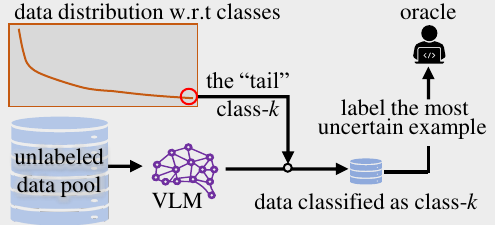}
  \hfill
  % \vspace{-0.5mm}
  \caption{\small
    {\bf Tail First Sampling (TFS)} utilizes the data distribution computed in the retrieved data to sample unlabeled data for the most under-represented class.
    Specifically, it first identifies this class over the current set of labeled data (including retrieved examples).
    Then, it finds the unlabeled examples classified as this classes.
    From this set, it selects the example that has the highest uncertainty of its prediction, e.g., the largest entropy.
    } 
\label{fig:TFS_workflow}
\end{SCfigure}
}

\subsection{Tail First Sampling for Active Learning}
\label{ssec:TFS}

\cref{fig:long-tail-distribution} demonstrates imbalanced or long-tailed distributions of retrieved data w.r.t the classes concerned by different downstream tasks (instantiated by different benchmarks).
Importantly, the imbalanced distribution of retrieved data implies how the pretrained VLM is biased for these classes.
Furthermore, as VLM's pretraining data is sampled from the real world,
the unlabeled task-specific data likely follows similar imbalanced distributions.
Motivated by these, we propose \emph{Tail First Sampling (TFS)}, which prioritizes sampling unlabeled examples for underrepresented-classes to label in AL.

In each round of AL,
TFS runs the VLM, adapted in the previous round, to produce pseudo-labels for unlabeled examples.
Over the examples that are pseudo-labeled as the ``tail classes'' (e.g., the most under-represented class),
TFS ranks them w.r.t prediction uncertainty computed by entropy.
It returns the top-ranked example for labeling by an oracle (e.g., human annotator).
The new labeled example is added to labeled data pool to adapt VLM in this AL round.
% \cref{fig:TFS_workflow} illustrates the workflow of TFS,
\cref{fig:TFS_workflow} summarizes the workflow of TFS;
\Cref{alg:tfs} contains pseudo code for TFS.
Next, we describe different VLM adaptation approaches.

{\em Remark.}
Our TFS selects rare-class data based on predicted pseudo labels, which can be 
error-prone.
Prior work finds that pseudo labels can be imbalanced even on class-balanced data~\cite{wang2022debiased, bang2024active}.
Hence, one may question whether relying on pseudo labels to sample hypothetical tail-class data is effective in AL.
Our experiments empirically demonstrate that TFS works well (\Cref{tab:benchmarking-results}).
We conjecture that regardless whether the pseudo labels for rare classes are correct or wrong,
the corresponding examples are valuable hard ones over which the (adapted) model makes more incorrect predictions.
As a result, including them in the next round for AL brings more performance gains.

\subsection{VLM Adaptation Approaches in Active Learning}
\label{ssec:adaptation}
\begin{figure*}[t]
  \centering
  \small
  \includegraphics[width=0.99\linewidth, clip=true, trim = 0mm 0mm 0mm 0mm]{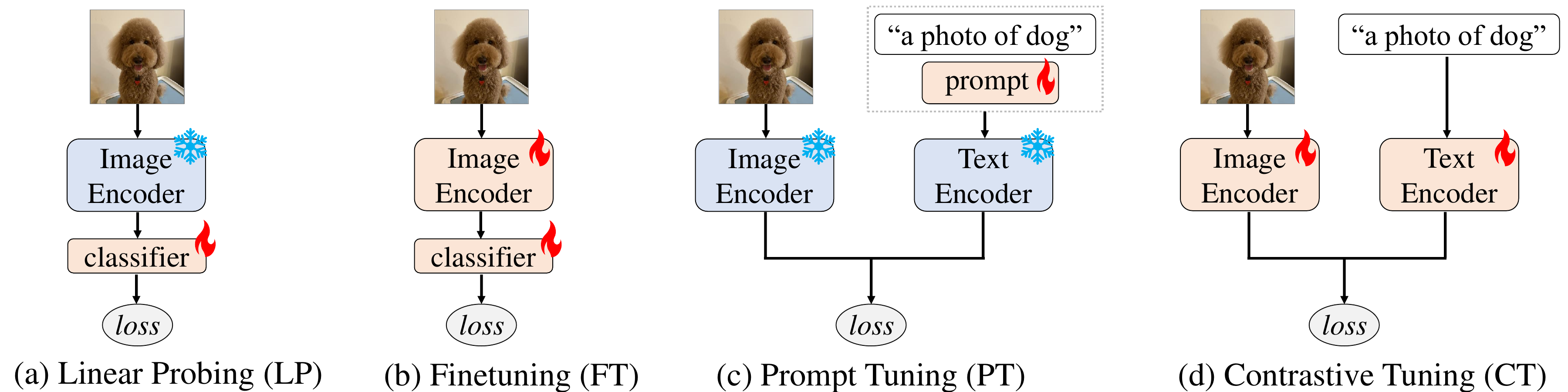}
  \vspace{-1mm}
  \caption{\small
  {\bf Conceptual comparison between four different adaptation approaches.}
  (a) Linear Probing (LP) learns a linear classifier on top of the frozen  pretrained visual encoder of a VLM. 
  (b) Typical finetuning (FT)
  learns to update parameters of the pretrained visual encoder.
  (c) Prompt Tuning (PT) learns parameters in the input space which are concatenated with text prompt over pretrained visual and text encoders. Note that PT is a \emph{de facto} approach to adapt VLM in the recent literature of active learning.
  (d) Contrastive Tuning (CT) learns to update both visual and text encoders using a contrastive loss.
  \Cref{tab:benchmarking-results}, \Cref{tab:CT_in_round-0} 
 and \cref{fig:compare_adaptation} demonstrate that CT resoundingly outperforms other adaptations approaches.
  }
  \vspace{-4mm}
  \label{fig:adaptation_methods}
\end{figure*}

Although contemporary AL methods commonly adopt prompt tuning (PT)~\cite{bang2024active, safaei2024active} to adapt a VLM,
we rigorously evaluate more adaptation approaches, including linear probing (LP), finetuning (FT), and contrastive tuning (CT)~\cite{goyal2023flyp}.
\cref{fig:adaptation_methods} conceptually compares these approaches.
PT~\cite{zhou2022coop} trains a set of learnable contexts; LP trains only the linear layer; FT updates all of the parameters of the VLM's image encoder and learns a linear classifier; CT~\cite{goyal2023flyp} trains both VLM's image and text encoders with a contrastive loss which is used to pretrain VLM.
Our experiments demonstrate that CT resoundingly outperforms other adaptation approaches,
regardless of whether using RDA or not and what AL method is used (ref. \Cref{tab:CT_in_round-0}, \Cref{tab:different_adaptation_woRDA}, and \cref{fig:compare_adaptation}).

\section{Experiments}
\label{sec:experiments}

We conduct extensive experiments to validate the effectiveness of Retrieval-based Data Augmentation (RDA) and Tail First Sampling (TFS), and compare different adaptation approaches.
We reiterate that our final method ALOR adopts RDA, TFS and CT,
achieving significantly better performance than prior arts.
We start by describing datasets, the evaluation protocol, important implementation details, and compared methods.

{\bf Datasets.}
AL is widely studied through the lens of image classification.
We follow this literature~\cite{bang2024active,safaei2024active} and use five established benchmark datasets in our work:
Semi-Aves~\cite{su2021semi_aves}, Food101~\cite{food}, FGVC-Aircraft~\cite{maji2013aircraft}, Stanford Cars~\cite{cars}, and OxfordPets~\cite{parkhi2012pets}. 
These datasets are publicly available for non-commercial research and educational purposes.
They have been widely used in broad domains such as active learning~\cite{bang2024active}, few-shot learning~\cite{Liu_2023_react, liu2025few} and zero-shot learning~\cite{parashar2024neglected, NEURIPS2023_NeuralPriming}.
We refer the reader to the supplement (\Cref{sec:datasets_suppl}) for details of these datasets.
In each dataset, we randomly sample labeled data from its official training set for Round-0 and use the rest training data as unlabeled data.
We evaluate methods on the official validation set.
Moreover, 
in our work, we use the open-source VLM OpenCLIP ViT-B/32 \cite{openclip} and its publicly-available pretraining dataset LAION-400M~\cite{laion400m} as the open data.
LAION-400M is open-source under CC-BY 4.0 License and is reported to be more suitable for retrieval-based tasks 
\cite{cherti2023reproducible}.

{
\setlength{\tabcolsep}{0.45em}
\begin{table*}[t]
\centering
\vspace{-3mm}
\caption{\small
\textbf{Results of different AL methods with (w/) vs. without (w/o) Retrieval-based Data Augmentation (RDA).}
We use prompt tuning (PT) as the adaptation approach,
which is the {\em de facto} approach in contemporary AL methods literature.
For each method on a specific dataset, we report the  accuracy (after the last round);
we run each method for three random runs.
Refer to \cref{fig:barchart-compare_rda} for per-round metrics of these methods.
Clearly, RDA boosts the performance for all the methods.
Moreover, our TFS outperforms the compared method. 
As TFS uses retrieved data (through RDA) to compute  class distribution and facilitate unlabeled data selection, it is not applicable without RDA.
Refer to \Cref{sec: Detail_results_suppl} for detailed results.
}
\vspace{-2mm}
\scalebox{0.845}
{
\begin{tabular}{lcccccccccccccccccccc}
\toprule
    &  \multicolumn{2}{c}{Semi-Aves} & \multicolumn{2}{c}{Aircraft} & \multicolumn{2}{c}{Stanford Cars} &\multicolumn{2}{c}{Food101} & \multicolumn{2}{c}{OxfordPets} & \multicolumn{2}{c}{\cellcolor{col3}\textbf{Average}}\\ 
    \textbf{Method}  & w/ & w/o  & w/ & w/o & w/ & w/o & w/ & w/o & w/ & w/o & \cellcolor{col3}w/ & \cellcolor{col3}w/o     \\
    \cmidrule(lr){1-1} \cmidrule(lr){2-3} \cmidrule(lr){4-5} \cmidrule(lr){6-7} \cmidrule(lr){8-9} \cmidrule(lr){10-11} \cmidrule(lr){12-13} 

    Entropy~\cite{holub2008entropy} & 51.02 & 32.98 & 33.04 & 23.73 & 82.57 & 76.44 & 68.06 & 57.79 & 80.96 & 73.62 & \cellcolor{col3}63.13 & \cellcolor{col3}52.91 \\
    Coreset~\cite{sener2018coreset} & 50.30 & 32.53 & 33.12 & 22.43 & 80.99 & 75.20 & 67.33 & 52.31 & 79.71 & 72.92 & \cellcolor{col3}62.29 & \cellcolor{col3}51.08 \\
    BADGE~\cite{Ash2020Badge} & 50.73 & 33.68 & 33.67 & 28.38 & 82.53 & 78.43 & 68.24 & 61.70 & 80.90 & 78.12 & \cellcolor{col3}63.21 & \cellcolor{col3}56.06 \\
    PCB + BADGE~\cite{bang2024active} & 50.74 & 34.29 & 34.00 & 28.18 & 82.68 & 78.04 & 68.08 & 63.22 & 80.80 & 76.10 & \cellcolor{col3}63.26 & \cellcolor{col3}55.97 \\
    ALFA-Mix~\cite{alfamix} & 50.36 & 33.62 & 33.14 & 28.55 & 79.77 & 76.81 & 67.82 & 66.31 & 80.76 & 75.90 & \cellcolor{col3}62.37 & \cellcolor{col3}56.24 \\
    LoGo~\cite{kim2023re} & 50.38 & 32.53 & 33.56 & 26.54 & 82.51 & 77.05 & 68.32 & 63.05 & 81.17 & 75.18 & \cellcolor{col3}63.19 & \cellcolor{col3}54.87 \\
    {\bf TFS}  & 51.45 & n/a & 34.19 & n/a & 82.52 & n/a & 68.34 & n/a & 81.12 & n/a & \cellcolor{col3}\textbf{63.52} & \cellcolor{col3}n/a \\

\bottomrule

\end{tabular}
}
\vspace{-3mm}
\label{tab:table_Comparison_RDA}
\end{table*}
}

\begin{figure*}[t]
  \centering
  \small
  \includegraphics[width=0.99\linewidth, clip=true, trim = 0mm 0mm 0mm 0mm]{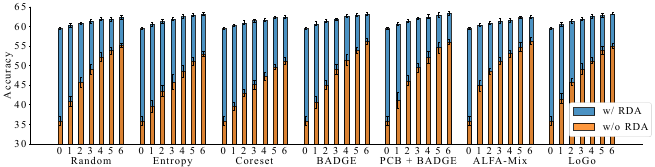}
  \vspace{-1mm}
  \caption{\small
  {\bf RDA significantly boosts active learning methods.}
  For each method in each round,
  we report the averaged accuracy and standard deviation over three random runs across five datasets.
  In Round-0,
  methods with RDA achieve 1.7$\times$ higher accuracy than without!
  In the last round (Round-6), 
  RDA helps each method obtain $>$8\% accuracy gains.
  }
\label{fig:barchart-compare_rda}
\vspace{-6mm}
\end{figure*}

{\bf Benchmarking Protocol.}
Following prior work~\cite{bang2024active}, we set 7 rounds (from 0 through 6) for AL.
Suppose there are $K$ classes,
Round-0 provides an initial set of $K$ labeled examples that each class has exactly one labeled image. 
The initial set allows VLM adaptation in the beginning of AL.
In a subsequent round, an AL method selects $K$ unlabeled examples from the unlabeled data pool,
sends them to an oracle for labeling, and adapts the VLM over all the labeled data.
For each AL method on a dataset, we run it three times with different random seeds and report the mean accuracy (with standard deviation) and macro-F1 score as evaluation metrics~\cite{bang2024active, li2013active, xu2018using}.
To simplify the comparisons,
we report summary numbers as the averaged metrics over all the three random runs, all the seven rounds, all the datasets or all the compared AL methods, depeneding on the context.
For fair comparison between methods when using RDA,
we use the same set of retrieved data.

{\bf Implementation Details.}
For RDA in Round-0, 
we retrieve data using string matching, finding pretraining texts that contain class names in the VLM's pretraining dataset.
We cap per-class retrieved examples to be no more than 500,
removing the retrieved images that are least relevant to the concerned class names measured by cosine similarity on features computed by the pretrained VLM.
In \Cref{sec:RDA_suppl} of the Supplement, we study the performance of training on different numbers of retrieved data and capping strategies (for balancing retrieved data). Results show that capping per-class retrieved examples by 500 not only performs the best (owing to the more balanced distribution) but also reduces compute cost (owing to the removal of abundant data). 
We run each method on a single NVIDIA 4090 GPU.
For VLM adaptation, we use a lower learning rate (1e-6) for the backbone and a higher learning rate (1e-4) for the classifier.
For CT, we update the temperature with a learning rate 1e-4.
In each round, we adapt the VLM for 50 epochs for each adaptation approach,
using the AdamW optimizer, batch size 32, and a weight decay 1e-2.
These hyperparameters are adopted by prior works~\cite{parashar2024neglected,lin2023multimodality,liu2025few} (see more details in \Cref{sec:hyperparams_suppl} of the Supplement).

{\bf Compared Methods.} 
We compare multiple representative and state-of-the-art methods of active learning.
\emph{Random} is a baseline that randomly selects unlabeled images in each round for labeling.
\emph{Entropy} selects unlabeled examples that have the largest entropy~\cite{holub2008entropy}, i.e., those that have the most uncertain predictions.
\emph{Coreset} select diverse unlabeled examples in the feature space computed with the VLM~\cite{sener2018coreset}.
\emph{BADGE} uses $k$-means clustering in the gradient space to select unlabeled examples that are diverse and have uncertain predictions by the VLM~\cite{Ash2020Badge}.
\emph{Pseudo-Class Balance (PCB)} selects unlabeled examples by balancing the distribution w.r.t pseudo-labels predicted by VLM~\cite{bang2024active}. It can be applies jointly with existing AL methods. We evaluate PCB + BADGE as this is reported to perform the best among compared methods in~\cite{bang2024active}.
\emph{ALFA-Mix} analyzes the neighborhood surrounding an unlabeled example by interpolating its features with those of previously labeled ones for unlabeled data selection~\cite{sener2018coreset}.
\emph{LoGo} first performs $k$-means clustering in the gradient space, then uses the EM algorithm to partition the clustering boundaries using the labeled examples from the previous round, followed by cluster sampling~\cite{kim2023re}.
While these methods use prompt tuning (PT) to adapt VLM,
we further compare other adaptation approaches (e.g., LP, FT, and CT), revealing that CT consistently performs the best.

\begin{table*}[t]
\centering
\small
\caption{\small
\textbf{Benchmarking results w.r.t accuracy and F1} after round-6 on each dataset.
Following the literature~\cite{bang2024active,safaei2024active},
we use prompt tuning (PT) in each AL method to adapt VLM.
We improve all the compared AL methods by applying RDA (\Cref{tab:table_Comparison_RDA}).
On each dataset, we bold and underline the best and second best numbers, respectively.
We also compare our TFS by using different adaptation approaches other than PT, including LP, FT, and CT.
Results show that our TFS outperforms existing AL methods when using PT for VLM adaptation.
Moreover, as \emph{our final ALOR method},  ``TFS w/ CT'' (with RDA) significantly boosts performance, achieving $\sim$7 points gains (in both accuracy and macro F1 averaged on the five datasets) over ``TFS w/ PT'' and existing AL methods.
Refer to Fig.~\ref{fig:compare_al} for per-round metrics of these methods and \Cref{sec: Detail_results_suppl} in the Supplement for more details.
}
\vspace{-2mm}
\setlength{\tabcolsep}{0.4em}
\scalebox{0.86}
{
\begin{tabular}{lcccccccccccccc}
    \toprule
        &
        & \multicolumn{2}{c}{Semi-Aves} 
        & \multicolumn{2}{c}{Aircraft} 
        & \multicolumn{2}{c}{Stanford Cars} 
        & \multicolumn{2}{c}{Food101}  
        & \multicolumn{2}{c}{OxfordPets} 
        & \multicolumn{2}{c}{\cellcolor{col3}\textbf{Average}} \\
        
        \textbf{Method} & \textcolor{gray}{venue} & Acc &  F1  & Acc &  F1  & Acc &  F1  & Acc &  F1 & Acc &  F1 & \cellcolor{col3}Acc & \cellcolor{col3}F1 \\
        
        \cmidrule(lr){1-1}
        \cmidrule(lr){2-2}
        \cmidrule(lr){3-4} 
        \cmidrule(lr){5-6}
        \cmidrule(lr){7-8} 
        \cmidrule(lr){9-10} 
        \cmidrule(lr){11-12} 
        \cmidrule(lr){13-14} 
        
        zero-shot~\cite{openclip} & \textcolor{gray}{CVPR'23} & 8.78 & 4.65 & 18.39 & 16.67 & 8.54 & 56.04 & 78.89 & 88.91 & 88.91 & 88.76 & \cellcolor{col3}50.69 & \cellcolor{col3}49.00 \\
        
        \midrule
        
        Random & & 50.09 & 49.57 & 33.94 & 32.06 & 79.95 & 79.09 & 67.58 & 66.88 & 79.92 & 79.51 & \cellcolor{col3}62.30 & \cellcolor{col3}61.42 \\
        
        Entropy~\cite{holub2008entropy}  & \textcolor{gray}{CVPR'08} & 51.02 & 50.50 & 33.04 & 30.95 & 82.57 & 81.91 & 68.06 & 67.25 & 80.96 & 80.67 & \cellcolor{col3}63.13 & \cellcolor{col3}62.26 \\
        CoreSet~\cite{sener2018coreset}  & \textcolor{gray}{ICLR'18} & 50.30 & 49.57 & 33.12 & 30.99 & 80.99 & 80.16 & 67.33 & 66.50 & 79.71 & 79.13 & \cellcolor{col3}62.29 & \cellcolor{col3}61.27 \\
        BADGE~\cite{Ash2020Badge} & \textcolor{gray}{ICLR'20} & 50.73 & 50.18 & 33.67 & 31.60 & 82.53 & 81.94 & 68.24 & 67.48 & 80.90 & 80.60 & \cellcolor{col3}63.21 & \cellcolor{col3}62.36 \\
        PCB + BADGE~\cite{bang2024active} & \textcolor{gray}{CVPR'24} & 50.74 & 50.21 & 34.00 & 32.22 & 82.68 & 82.20 & 68.08 & 67.33 & 80.80 & 80.45 & \cellcolor{col3}63.26 & \cellcolor{col3}62.48 \\
        ALFA-Mix~\cite{alfamix} & \textcolor{gray}{CVPR'22} & 50.36 & 49.91 & 33.14 & 31.27 & 79.77 & 78.81 & 67.82 & 67.18 & 80.76 & 80.48 & \cellcolor{col3}62.37 & \cellcolor{col3}61.53 \\
        LoGo~\cite{kim2023re} & \textcolor{gray}{CVPR'23} & 50.38 & 49.73 & 33.56 & 31.42 & 82.51 & 81.99 & 68.32 & 67.80 & 81.17 & 80.92 & \cellcolor{col3}63.19 & \cellcolor{col3}62.37 \\
        
        \midrule 
        
        {\bf TFS w/ PT}  &
        \textcolor{gray}{\bf ours} & 51.45 & 51.12 & 34.19 & 32.36 & 82.52 & 82.01 & 68.34 & 67.54 & 81.12 & 80.82 & \cellcolor{col3}63.52 & \cellcolor{col3}62.77 \\
    
        {\bf TFS w/ LP}  & 
        \textcolor{gray}{\bf ours}  & 51.41 & 51.09 & 34.88 & 33.49 & 80.96 & 80.00 & \textbf{75.77} & \textbf{75.54} & \textbf{87.64} & \textbf{87.57} & \cellcolor{col3}66.13 & \cellcolor{col3}65.54 \\
    
        {\bf TFS w/ FT}  & \textcolor{gray}{\bf ours}  & \underline{55.37} & \underline{54.94} & \underline{49.89} & \underline{48.47} & \underline{84.38} & \underline{83.61} & \underline{73.09} & \underline{72.41} & 85.04 & 84.74 & \cellcolor{col3}\underline{69.55} & \cellcolor{col3}\underline{68.83} \\
            
        {\bf TFS w/ CT}  &  
        \textcolor{gray}{\bf ours} & \textbf{56.95} & \textbf{56.90} & \textbf{50.84} & \textbf{49.77} & \textbf{85.75} & \textbf{85.47} & 72.86 & 72.33 & \underline{85.70} & \underline{85.44} & \cellcolor{col3}\textbf{70.42} & \cellcolor{col3}\textbf{69.98} \\
        
    \bottomrule
\end{tabular}
}
\label{tab:benchmarking-results} 
\end{table*}%

\begin{figure}[t]
    \begin{minipage}[b]{0.49\textwidth}
            \includegraphics[width=0.99\textwidth,
            clip=true,trim = 0mm 0mm 0mm 0mm]{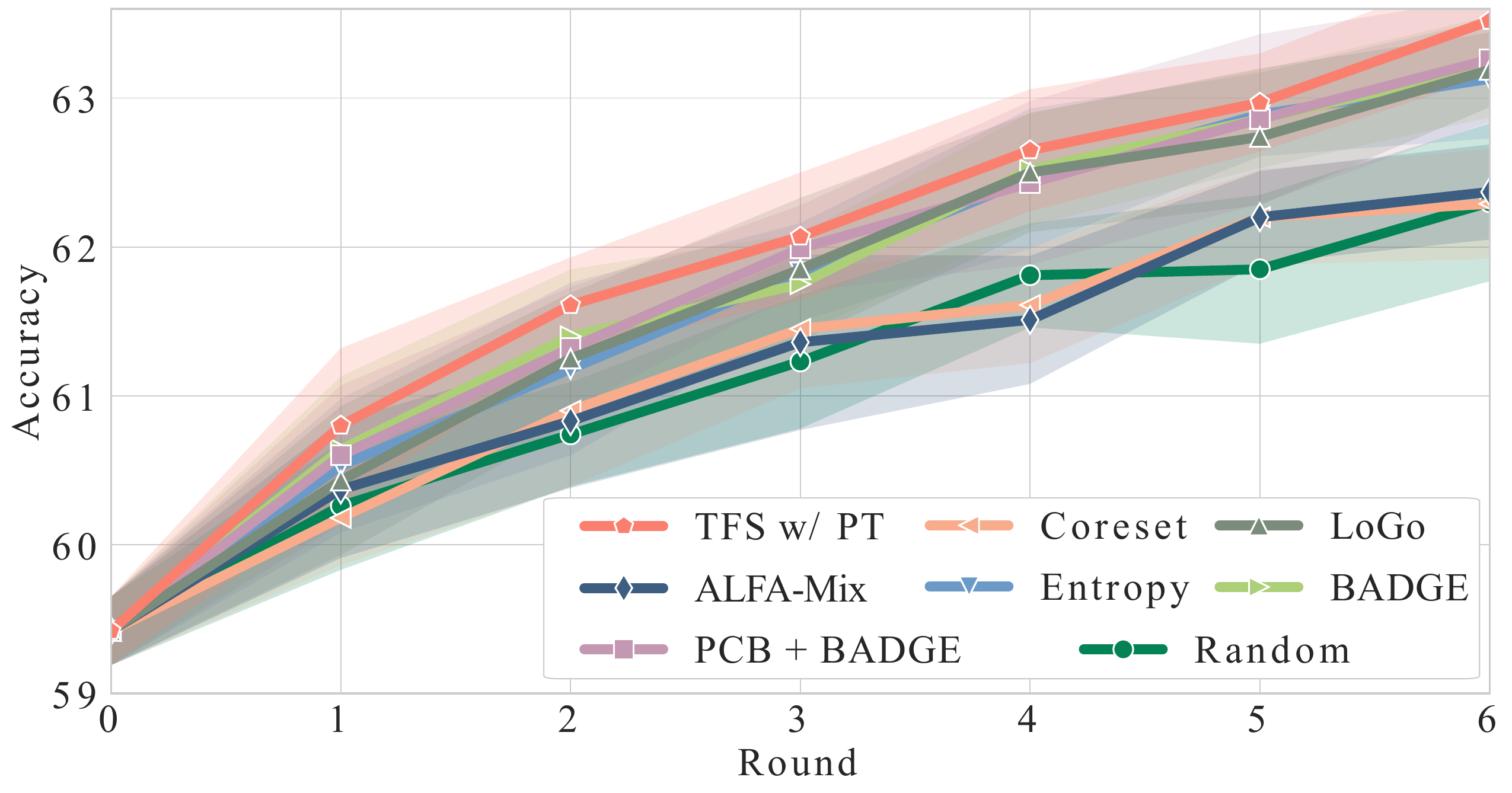}
    \end{minipage}%
    \hspace{\fill}%
    \begin{minipage}[b]{0.49\linewidth}
        \caption{\small
        {\bf Comparison of different AL methods w.r.t per-round accuracy.}
        For each method in each round,
        we report its averaged accuracy with standard deviation (reflected by the shallow) over three random runs across five different datasets.
        All methods start from the same pretrained VLM and adopt RDA and Prompt Tuning (PT).
        Results show that our \textcolor{fig3_tfs}{TFS} consistently outperforms the compared methods. 
        Refer to \Cref{tab:benchmarking-results} for results on each dataset.
        }
    \label{fig:compare_al}
    \end{minipage}
    \vspace{-6mm}
\end{figure}

\begin{figure*}[t]
  \centering
  \small
  \includegraphics[width=1\linewidth, clip=true, trim = 0mm 0mm 0mm 0mm]{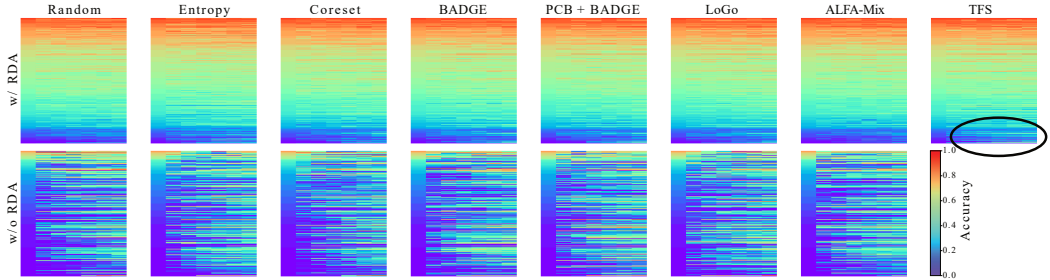}
  \vspace{-6mm}
  \caption{\small
  {\bf Visualization of per-round ($x$-axis) per-class ($y$-axis) accuracies on semi-Aves benchmark}.
  For each method, we sort its per-class accuracies in round-0 and track the accuracies  over time. 
  Results of different methods with and without RDA are shown in the top and bottom rows, respectively.
  When adopting RDA, all AL methods suffer from learning for under-represented classes, which have limited retrieved data compared to common classes.
  However, it is worth noting that our TFS can quickly improve on the under-represented classes as it prioritizes sampling data for them (see the quick improvements of under-represented classes pointed by the black circle.
  Without RDA, all methods do not have notable patterns in accuracy change although they yield better accuracies than previous rounds.
  This suggests the benefit of RDA that indicates how to sample data to improve on underrepresented classes and eventually enhance overall accuracy.
  }
  \vspace{-0mm}
  \label{fig:visual_comparison_performance}
\end{figure*}

\subsection{Benchmarking Results}

{\bf RDA significantly enhances existing AL methods.}
Table~\ref{tab:table_Comparison_RDA} compares the performance of different methods with-\emph{versus}-without RDA.
Clearly,
RDA, retrieving abundant data from the VLM's pretraining set for augmenting labeled task-specific data,
significantly enhances existing AL methods.
Furthermore,
\cref{fig:barchart-compare_rda} 
presents per-round accuracy for each method and compares the results between with-vs-without RDA.
The barcharts reconrfirm the advantages of using RDA to improve AL.
Importantly,
it is worth noting that, in Round-0, methods with RDA achieve 1.7$\times$ higher accuracy than their original versions that do not adopt RDA!

{\bf TFS outperforms previous AL methods.}
\Cref{tab:benchmarking-results} compares previous methods with our proposed TFS on each benchmark dataset, reporting accuracy after the last round.
As existing methods adopt prompt tuning (PT) to adapt VLM,
we list the results of our TFS when using PT.
Importantly,
we present results of our TFS that utilizes different adaptation methods including LP, FT, and CT.
For convenient comparison, we plot per-round accuracy of each method, averaged over five benchmark datasets, in \cref{fig:compare_al}.
Results show that our TFS consistently outperforms the compared methods,
demonstrating the benefit of exploiting data distribution obtained by RDA for unlabeled data selection.
Furthermore, 
in \cref{fig:visual_comparison_performance},
we analyze how the performance of each AL method on each class across AL rounds, with vs. without RDA.
When adopting RDA, all AL methods suffer from learning for under-represented classes,
on which RDA is unable to fetch enough data compared to common classes.
Nevertheless, it is worth noting that our TFS can quickly improve on the under-represented classes, owing to its mechanism that prioritizes these classes in data sampling.
Comparatively, without RDA, all methods do not show remarkable biases and do not have notable patterns in accuracy evolution.
The results demonstrate the benefit of using RDA to suggest how to sample unlabeled in AL, explaining why our TFS performs better than others.

\begin{table}[t]
    \begin{minipage}[t]{0.43\textwidth}
    \centering
    \setlength{\tabcolsep}{0.5em}
    \scalebox{0.9}
    {
        \begin{tabular}{lccccc}
        \toprule
            & \multicolumn{2}{c}{w/ RDA}  & \multicolumn{2}{c}{w/o RDA} \\ 
            approach  & Acc &  F1  & Acc &  F1 \\
            \cmidrule(lr){1-1} \cmidrule(lr){2-3} \cmidrule(lr){4-5} 
            PT & 47.31 & 46.22 & 14.14 & 12.91 \\
            LP & 47.90 & 46.64 & 15.53 & 14.57 \\
            FT & 51.50 & 50.20 & 16.33 & 14.21 \\
            CT & \bf 52.96 & \bf 52.40 & \bf 16.97 & \bf 15.28 \\
        \bottomrule
        \end{tabular}
    }
    \end{minipage}
    \hfill
    \begin{minipage}[t]{0.57\linewidth}
    \vspace{-16mm}
        \caption{\small
        \textbf{Contrastive Tuning (CT) outperforms other adaptation approaches}.
        We perform different adaptation approaches in Round-0,
        when there is the least number of labeled task-specific data, before AL methods are used.
        Results are on the challenging semi-Aves dataset.
        Somewhat surprisingly, CT performs the best, regardless of whether using RDA or not, significantly better than PT, which is commonly used in existing AL methods.
        }
    \label{tab:CT_in_round-0} 
    \end{minipage}
    \vspace{-4mm}
\end{table}

\begin{table*}[t]
\centering
\small
\caption{\small
\textbf{Comparison of different adaptation approaches} including Prompt Tuning (PT),
Linear Probing (LP),
Finetuning (FT),
and Contrastive Tuning (CT).
We run each combination of adaptation approach and AL method (without RDA) on the challenging Semi-Aves dataset for three random runs,
and report the mean accuracy after Round-6.
CT consistently outperforms other adaptation approaches regardless of what AL methods are used.
See \cref{fig:compare_adaptation} for more analysis.
}
\vspace{-2mm}
\setlength{\tabcolsep}{0.4em}
\scalebox{0.795}
{
\begin{tabular}{lcccccccccccccccccc}
\toprule
    &  \multicolumn{2}{c}{Random} 
    & \multicolumn{2}{c}{Entropy} 
    & \multicolumn{2}{c}{Coreset} 
    & \multicolumn{2}{c}{BADGE}  
    & \multicolumn{2}{c}{PCB + BADGE} 
    & \multicolumn{2}{c}{LoGo} 
    & \multicolumn{2}{c}{ALFA-Mix} 
    & \multicolumn{2}{c}{\cellcolor{col3}\textbf{Average}}
    \\
    \textbf{Method} & Acc &  F1  & Acc &  F1  & Acc &  F1  & Acc &  F1 & Acc &  F1 & Acc &  F1 & Acc &  F1 & \cellcolor{col3}Acc & \cellcolor{col3}F1 \\
    \cmidrule(lr){1-1}
    \cmidrule(lr){2-3}
    \cmidrule(lr){4-5} 
    \cmidrule(lr){6-7}
    \cmidrule(lr){8-9} 
    \cmidrule(lr){10-11} 
    \cmidrule(lr){12-13} 
    \cmidrule(lr){14-15} 
    \cmidrule(lr){16-17}

    {\bf PT} & 33.67 & 32.59 & 32.98 & 32.54 & 32.53 & 31.06 & 34.35 & 33.41 & 34.26 & 33.39 & 32.51 & 30.84 & 33.46 & 32.31 & \cellcolor{col3}33.39 & \cellcolor{col3}32.31 \\

    {\bf LP} & 34.19 & 32.89 & 35.40 & 34.38 & 33.61 & 31.73 & 35.19 & 34.10 & 34.86 & 34.02 & 35.28 & 34.16 & 33.98 & 32.88 & \cellcolor{col3}34.64 & \cellcolor{col3}33.45
    \\

    {\bf FT} & 32.59 & 30.50 & 33.40 & 32.21 & 31.80 & 29.93 & 33.24 & 31.22 & 34.84 & 33.45 & 33.45 & 31.74 & 33.43 & 32.12 & \cellcolor{col3}33.25 & \cellcolor{col3}31.60
    \\
        
    {\bf CT}  & 36.54 & 35.29 & 35.75 & 34.59 & 34.93 & 33.89 & 36.91 & 35.63 & 36.58 & 35.56 & 36.47 & 35.50 & 35.75 & 34.58 & \cellcolor{col3}{\textbf{36.13}} & \cellcolor{col3}{\textbf{35.01}} 
    \\
\bottomrule
\end{tabular}
}
\vspace{-1mm}
\label{tab:different_adaptation_woRDA} 
\end{table*}%

\begin{figure}[t]
    \begin{minipage}[b]{0.49\textwidth}
        \includegraphics[width = 1\textwidth, clip=true,trim = 0mm 0mm 0mm 0mm]{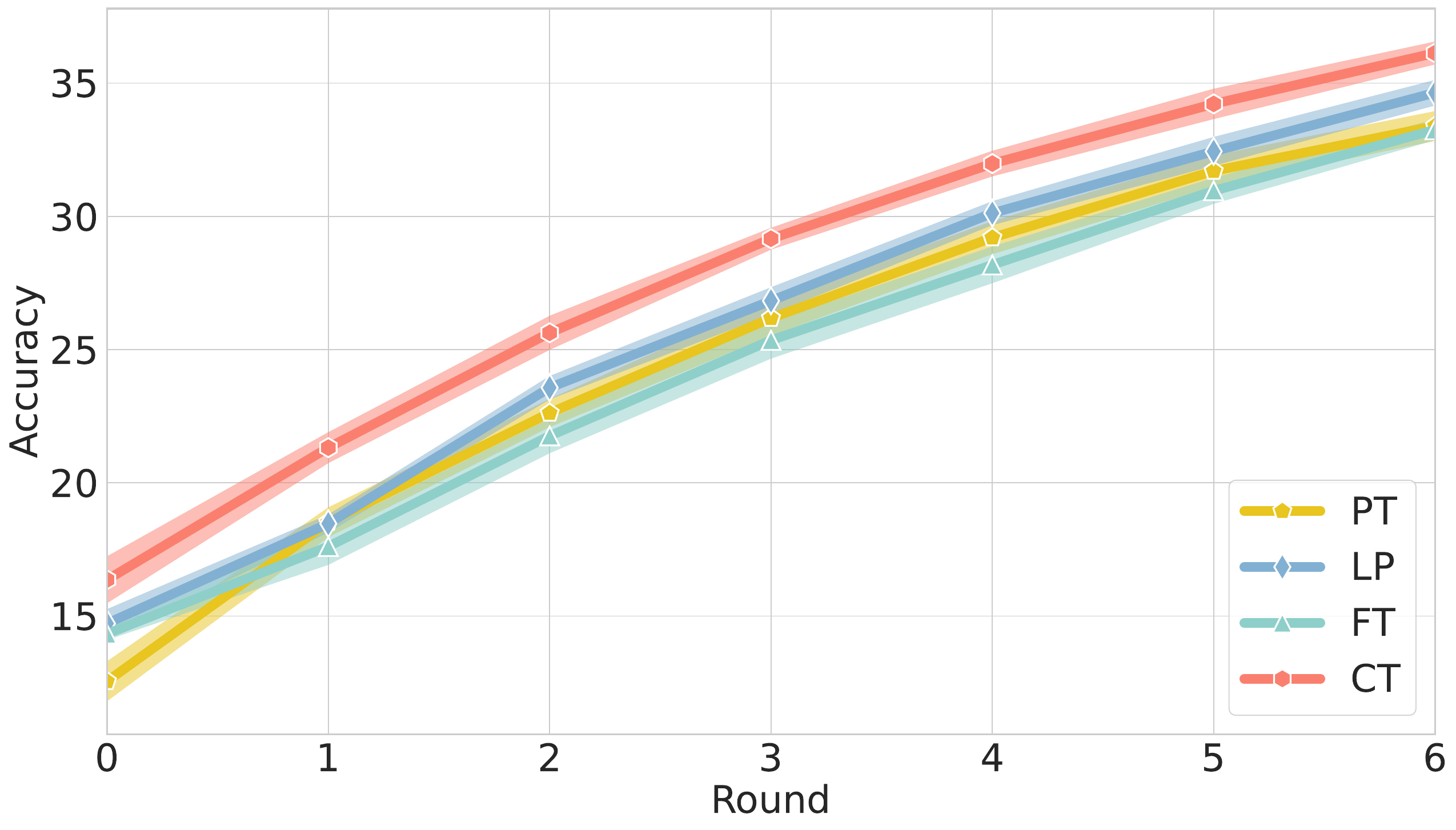}
    \end{minipage}%
    \hspace{\fill}%
    \begin{minipage}[b]{0.49\linewidth}
        \caption{\small
        {\bf CT consistently outperforms other adaptation approaches} (PT, LP, and FT). 
        To demonstrate this, we apply each adaptation approach to the compared AL methods listed in \Cref{tab:different_adaptation_woRDA},
        for three random runs,
        on the five benchmark datasets.
        Following the conventional setup, we do not adopt RDA.
        For each adaptation approach in each round,
        we report its mean accuracy and standard deviation over all AL methods, random runs, and datasets.
        CT outperforms other adaptation approaches.
        Refer to \Cref{tab:different_adaptation_woRDA} for more results.
        }
    \label{fig:compare_adaptation}
    \end{minipage}
    \vspace{-4.5mm}
\end{figure}

{\bf It is better to finetune VLM.}
We compare different VLM adaptation methods including Prompt Tuning (PT)~\cite{zhou2022coop} , Linear Probing (PL)~\cite{radford2021learning}, Finetuning (FT)~\cite{liu2025few}, and Contrastive Tuning (CT)~\cite{goyal2023flyp}.
\Cref{tab:CT_in_round-0} shows that CT performs the best even in the low-data regime (i.e., when not using RDA).
\Cref{tab:different_adaptation_woRDA} shows that CT consistently outperforms other adaptation approaches regardless of what AL methods are used,
significantly better than PT, which has been exclusively studied in the recent AL literature.
Moreover, \cref{fig:compare_adaptation} shows that CT outperforms other adaptation approaches in each round over all AL methods.
All the results challenge the current practice that uses PT for AL research.
Importantly,
our final method ALOR, by utilizing CT, boosts AL performance as shown in \Cref{tab:benchmarking-results} .

{\bf Our final method ALOR performs the best.}
Based on the previous results, we derive the final method ALOR by combining  CT, RDA, and TFS.
\Cref{tab:benchmarking-results} compares ALOR and others (with RDA for enhancement).
Results show that ALOR significantly outperforms all the compared methods, e.g., it achieves +$7.2$ accuracy gains over the previous methods PCB~\cite{bang2024active} and LoGo~\cite{kim2023re}.

\section{Discussions}
\label{sec:discussions}

{\bf Societal Impacts.}
Active learning (AL) is motivated by the desire to reduce annotation cost and thus has broader impacts on real-world applications and interdisciplinary research that adopt machine learning solutions. 
However, contemporary literature of  AL has insufficiently considered fairness and biases in the selected unlabeled data and the utilization of pretrained foundation models.
Our proposed AL method is not an exception.
Moreover, the retrieved examples through our retrieval data augmentation follow imbalanced and long-tailed distributions w.r.t class labels.
They might also be imbalanced distributed w.r.t other attributes and could cause unnoticed bias issues.
Lastly, although we follow the literature to use standard benchmark datasets in experiments,
we note that AL is expected to help data annotation in highly-specialized applications, e.g., annotating biological or medical specimens which require domain expertise to label data. Our work does not have a chance to delve into more real-world applications.

{\bf Limitations and Future Work.}
We note some limitations in our work and point out possible future directions. 
First, although adopting Retrieval-based Data Augmentation (RDA) significantly boosts active learning performance,
the retrieved data has domain gaps with the task-specific data.
It is worth exploring how to mitigate the domain gap in future work.
Second, our Tail First Sampling (TFS) relies on pseudo labels to select rare-class unlabeled examples; yet, the pseudo labels can be biased due to the inherent bias of pretrained VLM. Future work should delve into this issue.
Moreover, while prioritizing tail classes in data selection empirically enhances active learning,
TFS can be trapped into tail classes if the budget is too small and imbalance ratio is too high. This would make TFS constantly fetch data predicted as a specific rare class.
Future work can improve TFS by incorporating diversity constraints.
Lastly, to adapt VLM in active learning, although Contrastive Tuning (CT) significantly outperforms other VLM adaptation approaches,
it may still be prone to overfitting. Hence, future work can improve VLM adaptation approaches.

\section{Conclusions}
We study active learning (AL) by embracing open sources, including a pretrained Vision-Language Model (VLM) and its pretraining dataset, which are open-source and publicly available.
To note, existing AL methods have exploited a VLM but yet to leverage its pretraining data.
We show that retrieval-based data augmentation (RDA), by retrieving data relevant to a downstream task for VLM adaptation, significantly enhances AL on this task.
Moreover, over the retrieved data, we estimate data distribution w.r.t class labels concerned by the task.
This distribution tends to be imbalanced or long-tailed, implying how the pretrained VLM is biased and how task-specific data is similarly imbalanced distributed.
This insight motivates our rather simple yet effective unlabeled data sampling strategy, Tail First Sampling (TFS), which prioritizes sampling data for the rarest classes based on the predicted labels.
Through extensive experiments on five benchmark datasets, we show that TFS outperforms existing AL methods.
Lastly, we rigorously evaluate different adaptation approaches including Prompt Tuning, Linear Probing, Finetuning and Contrastive Tuning, with the first being widely adopted in the contemporary literature of active learning.
Comprehensive results demonstrate that CT resoundingly outperforms others.

% \section*{Acknowledgements}
% This work was supported by FDCT (0067/2024/ITP2), 
% the University of Macau (SRG2023-00044-FST), and the Institute of Collaborative Innovation.
% Authors thank Hanxin Wang for valuable discussions.

{
\small
\bibliographystyle{plainnat}
\bibliography{main}
}

\newpage

\clearpage

\setcounter{page}{1}
% \maketitlesupplementary

\renewcommand{\thesection}{\Alph{section}}
\renewcommand{\theHsection}{\Alph{section}}
\setcounter{section}{0}

\begin{center}
{{\bf \Large  Active Learning via Vision-Language Model \\
Adaptation with Open Data} 
\\
\vspace{3mm}
\emph{\Large (Supplemental Document)}}
\end{center}

This document supports our main paper with detailed results and comprehensive analyses. The document is organized as follows: 

\begin{itemize}
\item {\bf Section \ref{sec:datasets_suppl}} summarizes bencnhmark datasets. 

\item {\bf Section \ref{sec:hyperparams_suppl}} describes 
more learning details and hyper-parameters.

\item {\bf \Cref{sec:RDA_suppl}} provides details of retrieval-based data augmentation.

\item {\bf Section \ref{sec:open_source_code_suppl}} 
introduces the open-source code, as a part of supplementary material.

\item {\bf Section \ref{sec: Detail_results_suppl}} provides detailed results on  each dataset with each random seed.

\end{itemize}

\section{Summary of Datasets}
\label{sec:datasets_suppl}

\begin{figure*}[ht]
  \centering
  \small
  \includegraphics[width=0.9\linewidth, clip=true, trim = 0mm 0mm 0mm 0mm]{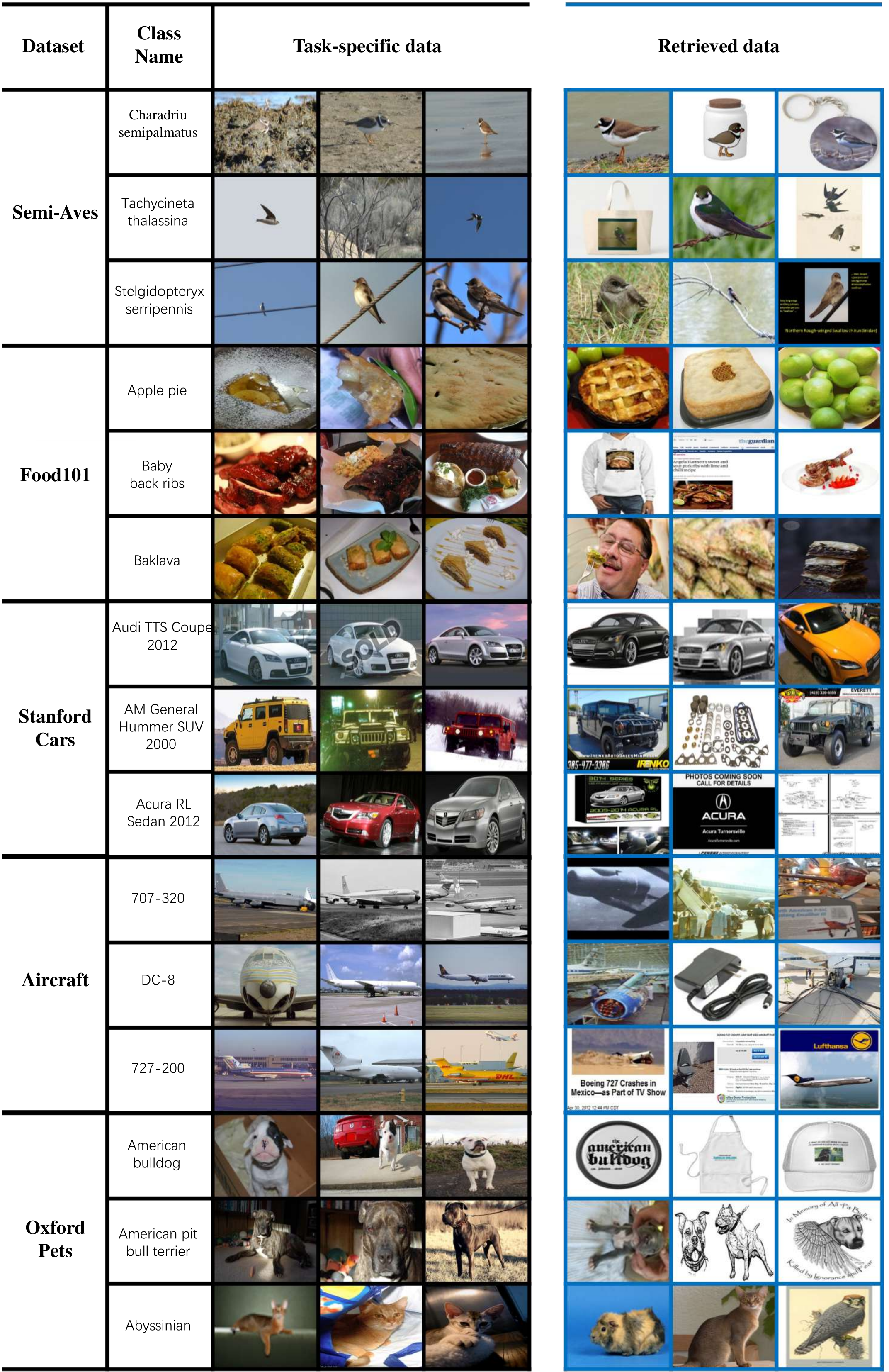}
  \vspace{-1mm}
  \caption{\small
  {\bf Examples of task-specific data and retrieved data.} 
  We showcase task-specific data from our benchmark datasets alongside their retrieved data from LAION-400M~\cite{laion400m}. Comparative analysis reveals significant domain discrepancies between task-specific data and retrieved data. For instance: A query for "727-200" (Aircraft dataset) retrieves an eBay product listing titled "BOEING 727 COCKPIT JUMP SEAT USED AIRCRAFT PART" (commercial marketplace context) rather than airplane imagery, as shown in the fifth example of Aircraft: 727-200; A search for "apple pie" (Food-101 dataset) returns images of raw baking ingredients instead of prepared desserts, exemplified by the final entry of Food101: Apple pie.
   }
\label{fig:data_detail_suppl}
\vspace{-3mm}
\end{figure*}

We summarize the five fine-grained datasets (OxfordPets~\cite{parkhi2012pets}, Semi-Aves~\cite{su2021semi_aves}, Aircraft~\cite{maji2013aircraft}, Food101~\cite{food} and StanfordCars~\cite{cars}) and LAION-400M~\cite{laion400m} retrieved data used in our experiments in Table~\ref{tab:datasets_suppl}, with visual examples provided in Figure~\ref{fig:data_detail_suppl}. These datasets operate under distinct licenses: OxfordPets is CC BY-SA 4.0 compliant for research use, while Semi-Aves, Aircraft, Food-101, and StanfordCars are restricted to non-commercial academic purposes; LAION-400M adopts the permissive CC-BY 4.0 license. Following established active learning protocols~\cite{bang2024active,safaei2024active}, we initialize training by randomly sampling labeled data from each dataset's training dataset and evaluate models on held-out test splits.
 We provide dataset statistics (e.g., data splits) in the \Cref{tab:datasets_suppl}.

{
\setlength{\tabcolsep}{0.5em}
\begin{table*}[h]
\centering
\small
\caption{\small \textbf{Details of five fine-grained datasets in our work.} We provide comprehensive statistics for our five fine-grained benchmarks, detailing per-dataset image counts (training / validation / test splits), retrieved data counts, and class counts. The training set serves as our \emph{unlabeled pool} for active selection, excluding an initially labeled subset that remains fixed across experiments. To ensure statistical robustness, all active learning rounds are executed three times with distinct random seeds.
}
\label{tab:datasets_suppl}
% \vspace{-3mm}
\scalebox{1}{
\begin{tabular}{lccccc}
\toprule 
dataset & classes &training set & validation set & test set & retrieved data \\
\midrule
Semi-Aves~\cite{su2021semi_aves} & 200 & 3,959 & 2,000 & 8,000 & 50,520\\
Aircraft~\cite{maji2013aircraft} & 100 & 3,334 & 3,333 & 3,333 & 32,222 \\
OxfordPets~\cite{parkhi2012pets} &37 &2,944 & 736 & 3,669 & 18,500 \\
Food101~\cite{food} &101 &50,500 & 20,200 & 30,300 & 50,220 \\
StanfordCars~\cite{cars} &196 &6,509 & 1,635 & 8,041 & 40,924 \\
\bottomrule
\end{tabular}}
% \vspace{-3mm}
\end{table*}
}

\section{Hyper-Parameter Setting}
\label{sec:hyperparams_suppl}

In our work, when we finetune the VLM, we employ a differential learning rate strategy: 
 the backbone network is optimized with a lower learning rate (1e-6), while classifiers utilize a higher rate of  (1e-4). 
The temperature parameter is initialized to 0.07 and updated at a learning rate of 1e-4.
For experiments with RDA, models are trained for 50 epochs, whereas without RDA configurations extend to 200 epochs. Following~\cite{parashar2024neglected,lin2023multimodality,liu2025few}, 
we adopt the AdamW optimizer with a batch size of 32 and weight decay of 1e-2, switching to SGD (lr=0.002) for prompt learning tasks~\cite{zhou2022coop,bang2024active}. Active learning initialization involves sampling one image per class (total $K$ images, where $K$ equals class count) from each dataset's training split, with seeds fixed at 666, 777, and 888 for reproducibility. 
Subsequent rounds (1-6) actively select $K$ images per round.
Following PCB~\cite{bang2024active}, 
we first sampling 10\% of unlabeled data via BADGE and select $K$ samples from this set. We recommend allocating >35GB storage and >5GB GPU RAM for experiments.

\section{Details of Retrieving Open Data}
\label{sec:RDA_suppl}

In previous studies, Retrieval Data Augmentation (RDA) has been used in zero-shot learning~\cite{Liu_2023_react, parashar2024neglected, NEURIPS2023_NeuralPriming} and few-shot learning~\cite{liu2025few}, yet its effectiveness is constrained by the long-tailed distribution of retrieved data (\Cref{fig:long-tail-distribution}), which introduces learning instability.
Prior work~\cite{liu2025few} shows that capping the maximum number of per-class retrieved data helps improve the final accuracy when learning on the retrieved long-tailed distributed data.
We study how to cap in more detail.
In particular, we study the results of capping the top-X dominant classes such that these classes have no more than Y retrieved examples or no more than Y\% examples compared to the retrieved data in the corresponding classes.
\Cref{tab:semi_aves_retrived_size_suppl} lists the results,
showing that hard-capping per-class retrieved examples produces better results, owing to a more balanced distribution after capping despite fewer retained images of each class. Based on the results in \Cref{tab:semi_aves_retrived_size_suppl}, we filtered the retrieval data from LAION-400M~\cite{laion400m} utilized text-to-image (T2I) similarity~\cite{Liu_2023_react}, and the numbers of final retrieval images for each dataset are shown in \Cref{tab:datasets_suppl}. 

{
\begin{table*}[t]
\vspace{0mm}
\caption{\small
\textbf{A study of capping strategies in retrieval-based data augmentation (RDA).} 
We report results in round-0 on the Semi-Aves dataset.
On the upper panel, we vary the capping number, which specifies  the maximum number of per-class retrieved examples;
we also select the top-X dominant classes to cap by varying X to be 20, 50, 100, and 200.
Moreover, 
we study another capping strategy, which proportionally caps per-class images based on the total number of retrieval examples for the specific class.
Ration-based capping will mitigate but retain imbalanced distributions; the results are listed in the lower panel. Interestingly, comprehensive results show that capping by number, in a way of forcing data to be balanced distributed across classes, performs better than ratio-based capping which retains but mitigate imbalanced distributions. The results are obtained after 100 epochs.
Moreover, capping more classes achieves better performance although retaining less data (as highlighted by bold numbers).
}
\vspace{-2mm}
\scalebox{0.725}
{
\begin{tabular}{lcrrrrrrr@{}}
\toprule
   \rowcolor{lightgrey} & & \multicolumn{4}{c}{Top-X classes with the highest number} \\
   \rowcolor{lightgrey} \textbf{capping \#} & &  
    \multicolumn{1}{c}{20} & \multicolumn{1}{c}{50} & \multicolumn{1}{c}{100} & \multicolumn{1}{c}{200} \\ 
    \midrule
    \multirow{2}{*}{100} & Acc & 53.18 & 52.40 & 51.13 & 50.29 \\
    & \#imgs & 61,684 & 31,310 & 17,858 & 16,386 \\
    \midrule
    \multirow{2}{*}{300} & Acc & 54.03 & 54.30 & 54.24 & 54.24 \\
    & \#imgs & 65,684 & 41,310 & 37,151 & 37,151 \\
    \midrule
    \multirow{2}{*}{500} & Acc & 54.64 & 54.74 & \bf 55.43 & \bf 55.43 \\
    & \#imgs & 69,684 & 51,310 & 50,720 & 50,720 \\
    
    \bottomrule
    \end{tabular}
    }
\hfill
\vspace{0mm}
\scalebox{0.725} 
{
\begin{tabular}{lcrrrrrrr@{}}
\toprule    
   \rowcolor{lightgrey} & & \multicolumn{4}{c}{Top-X classes with the highest number} \\
    \rowcolor{lightgrey} \textbf{capping \%} & &  
    \multicolumn{1}{c}{20} & \multicolumn{1}{c}{50} & \multicolumn{1}{c}{100} & \multicolumn{1}{c}{200} \\ 
    \midrule
    \multirow{2}{*}{20\%} & Acc & 54.08 & 53.88 & 51.44 & 47.03 \\
    & \#imgs & 87,487 & 60,827 & 46,135 & 40,167 \\
    \midrule
    \multirow{2}{*}{50\%} & Acc & 54.96 & 54.88 & 53.38 & 50.38 \\
    & \#imgs & 129,113 & 112,434 & 103,221 & 99,418 \\
    \midrule
    \multirow{2}{*}{70\%} & Acc & 54.70 & 54.38 & 52.55 & 52.68 \\
    & \#imgs & 156,906 & 146,934 & 141,459 & 139,299 \\
\bottomrule

\end{tabular}
}
\label{tab:semi_aves_retrived_size_suppl} %
\vspace{-3mm}
\end{table*}
}

\section{Open-Source Code}
\label{sec:open_source_code_suppl}
Our code is included in the Supplementary Material.

{\bf Requirements}.
Running the code requires Python and PyTorch.
We also use the open-source code of OpenCLIP~\cite{openclip}.
For full reproducibility, detailed package specifications are provided in the {\tt requirements.txt} file. Below are the critical software versions used in this work:
\begin{itemize}
\item Python: v3.10.14
\item PyTorch: v2.4.0
\end{itemize}

{\bf License}.
We release open-source code under the MIT License to foster future research in this field.

{\bf Instructions.}
Detailed instructions for reproducing our experiments are provided in the following documentation files:
\begin{itemize}
\item
{\tt requirements.txt} lists all dependencies to set up the Conda environment. 
\item
{\tt DATASETS.md} specifies steps to download benchmark datasets, predefined data splits.
\item
{\tt README.md} guides users on executing the code for various adaptation and active learning (AL) methods described in the paper.
\end{itemize}

{\bf Demo.}
The accompanying Jupyter Notebook provides an implementation of our \emph{Active Learning with Open Resources (ALOR)} framework. In this demonstration, we fine-tune the VLM~\cite{openclip} with both retrieved data from a VLM's pretraining dataset (i.e., LAION-400M~\cite{laion400m}) and task-specific data (i.e., Semi-Aves~\cite{su2021semi_aves}). And we employ our TFS as the AL method, selecting the most informative unlabeled samples for annotation and incorporation into subsequent rounds.
\begin{itemize}
\item
{\tt ALOR\_demo.ipynb} 

Running this file finetune~\cite{goyal2023flyp} the OpenCLIP (ViT-B/32)~\cite{openclip} using both the initial set (Round 0) and retrieved data. The process then employs our TFS to actively select the most informative unlabeled samples from the remaining pool, which are subsequently incorporated into train set for Round 1.

\end{itemize}

\section{Detailed Results on Different Datasets}
\label{sec: Detail_results_suppl}
In ~\Cref{tab:five_datasets_rda_suppl}, we multiple AL methods~\cite{alfamix, Ash2020Badge, sener2018coreset, holub2008entropy, bang2024active,kim2023re} against our TFS across different adaptation strategies (Rounds 0-6) on five datasets, using OpenCLIP (ViT-B/32)~\cite{openclip} with RDA. Results show that our method TFS w/ CT consistently outperforms baselines, achieving state-of-the-art performance across all benchmarks.
We also show the outcomes of each round for five datasets using different AL methods without RDA in ~\Cref{tab:five_datasets_without_rda_suppl}. Clearly, RDA
boosts the performance for all the methods. As TFS uses retrieved data (through RDA) to estimate class distribution and facilitate unlabeled data selection, it is not
applicable without RDA. Notably, all AL methods share identical Round-0 performance since they initialize training on the same labeled subset.

{
\setlength{\tabcolsep}{0.3em}
\begin{table*}[t]
\centering
\caption{\small
\textbf{Comparative Analysis of TFS and AL methods with OpenCLIP (ViT-B/32).} We present a detailed comparison of TFS against established AL methods across five benchmarks using OpenCLIP (ViT-B/32). Results are averaged over three random runs. We highlight the best number of round-6 in \textbf{bold} and \underline{underline} the second number of round-6 for each dataset. The complete results for each seed showed in~\Cref{tab:five_datasets_rda_suppl_seed666,tab:five_datasets_rda_suppl_seed777,tab:five_datasets_rda_suppl_seed888}. 
}
\vspace{-2mm}
\scalebox{0.61}
{
\begin{tabular}{llccccccccccccccc}
\toprule
    &
    & \multicolumn{2}{c}{Round 0} 
    & \multicolumn{2}{c}{Round 1} 
    & \multicolumn{2}{c}{Round 2} 
    & \multicolumn{2}{c}{Round 3}  
    & \multicolumn{2}{c}{Round 4} 
    & \multicolumn{2}{c}{Round 5}
    & \multicolumn{2}{c}{Round 6}
    \\
    \cmidrule(lr){3-4} 
    \cmidrule(lr){5-6}
    \cmidrule(lr){7-8} 
    \cmidrule(lr){9-10} 
    \cmidrule(lr){11-12} 
    \cmidrule(lr){13-14} 
    \cmidrule(lr){15-16} 
    
    \textbf{Dataset} & \textbf{Method} & Acc & Macro F1  & Acc & Macro F1  & Acc & Macro F1  & Acc & Macro F1 & Acc & Macro F1 & Acc & Macro F1 & Acc & Macro F1 \\
    \cmidrule(lr){1-1}
    \cmidrule(lr){2-2}
    \cmidrule(lr){3-4} 
    \cmidrule(lr){5-6}
    \cmidrule(lr){7-8} 
    \cmidrule(lr){9-10} 
    \cmidrule(lr){11-12} 
    \cmidrule(lr){13-14} 
    \cmidrule(lr){15-16} 

    & Random  & 47.18 & 46.03 & 48.25 & 47.06 & 48.58 & 47.64 & 48.68 & 47.93 & 49.36 & 48.72 & 49.73 & 49.11 & 50.09 & 49.57 \\
    & Entropy~\cite{holub2008entropy}  & 47.18 & 46.03 & 48.08 & 47.04 & 48.65 & 47.74 & 49.48 & 48.86 & 49.78 & 48.82 & 50.76 & 50.27 & 51.02 & 50.50 \\
    & CoreSet~\cite{sener2018coreset}  & 47.18 & 46.03 & 48.19 & 46.95 & 48.17 & 46.86 & 49.15 & 48.14 & 49.72 & 48.79 & 50.09 & 49.25 & 50.30 & 49.57 \\
    & BADGE~\cite{Ash2020Badge}  & 47.18 & 46.03 & 48.25 & 47.24 & 48.96 & 48.01 & 49.55 & 48.89 & 50.03 & 49.14 & 50.30 & 49.68 & 50.73 & 50.18 \\
    & PCB + BADGE~\cite{bang2024active} & 47.18 & 46.03 & 48.55 & 47.50 & 48.82 & 47.85 & 49.02 & 48.20 & 49.97 & 49.17 & 50.48 & 50.00 & 50.74 & 50.21 \\
    \textbf{Semi-Aves} & ALFA-Mix~\cite{alfamix}  & 47.18 & 46.03 & 48.11 & 47.21 & 48.70 & 47.69 & 49.46 & 48.85 & 49.34 & 48.79 & 50.15 & 49.56 & 50.36 & 49.91 \\
    & LoGo~\cite{kim2023re}  & 47.18 & 46.03 & 48.19 & 47.23 & 48.87 & 48.10 & 49.10 & 48.26 & 49.53 & 48.73 & 50.06 & 49.31 & 50.38 & 49.73 \\
    \cmidrule(lr){2-16}
    
    & {\bf TFS w/ PT}  & 47.18 &	46.03 &	48.99 &	48.05 &	49.68 &	49.05 &	50.24 &	49.71 &	50.72 &	50.20 &	51.06 &	50.68 &	51.45 &	51.12 \\

    & {\bf TFS w/ LP}  & 47.84 &	46.58 &	48.84 &	48.00 &	49.81 &	49.14 &	50.27 &	49.81 &	50.79 &	50.32 &	51.44 &	51.05 &	51.41 &	51.09 \\

    & {\bf TFS w/ FT}  & 51.46 &	50.19 &	52.79 &	51.99 &	53.47 &	52.70 &	54.56 &	53.95 &	54.40 &	53.91 &	54.95 &	54.46 &	\underline{55.37} &	\underline{54.94} \\

    & {\bf TFS w/ CT}  & 53.01 &	52.35 &	54.56 &	54.33 &	55.10 &	55.03 &	55.73 &	55.69 &	56.12 &	56.03 &	56.37 &	56.32 &	\bf56.95 &	\bf56.90 \\
    
    \midrule 
    
    & Random  & 31.08 & 28.26 & 31.23 & 28.43 & 32.32 & 29.98 & 32.57 & 29.88 & 33.31 & 31.18 & 33.11 & 30.95 & 33.94 & 32.06 \\
    & Entropy~\cite{holub2008entropy} & 31.08 & 28.26 & 31.15 & 28.12 & 31.13 & 28.85 & 31.85 & 29.30 & 32.84 & 30.80 & 32.48 & 30.36 & 33.04 & 30.95 \\
    & CoreSet~\cite{sener2018coreset}  & 31.08 & 28.26 & 31.69 & 28.91 & 32.23 & 29.43 & 32.43 & 29.93 & 32.16 & 29.93 & 33.47 & 31.24 & 33.12 & 30.99 \\
    & BADGE~\cite{Ash2020Badge} & 31.08 & 28.26 & 31.32 & 28.42 & 31.72 & 29.21 & 32.28 & 29.86 & 33.05 & 30.76 & 32.82 & 30.75 & 33.67 & 31.60 \\
    & PCB + BADGE~\cite{bang2024active}  & 31.08 & 28.26 & 31.83 & 29.09 & 32.00 & 29.24 & 33.27 & 31.02 & 33.11 & 31.32 & 33.64 & 31.56 & 34.00 & 32.22 \\
    \textbf{Aircraft} & ALFA-Mix~\cite{alfamix}   & 31.08 & 28.26 & 31.82 & 29.34 & 32.16 & 29.37 & 32.51 & 29.67 & 32.22 & 29.78 & 33.56 & 31.30 & 33.14 & 31.27 \\
    & LoGo~\cite{kim2023re}  & 31.08 & 28.26 & 31.24 & 28.13 & 32.27 & 29.70 & 31.94 & 29.15 & 32.86 & 30.28 & 33.47 & 31.35 & 33.56 & 31.42 \\
    \cmidrule(lr){2-16}

    & {\bf TFS w/ PT}  & 31.08 &	28.26 &	32.38 &	29.68 &	32.86 &	30.24 &	33.61 &	31.44 &	33.60 &	31.75 &	34.22 &	32.14 &	34.19 &	32.36 \\

    & {\bf TFS w/ LP}  & 31.52 &	28.58 &	32.35 &	29.93 &	32.82 &	30.85 &	33.37 &	31.50 &	33.76 &	31.92 &	34.63 &	33.22 &	34.88 &	33.49 \\

    & {\bf TFS w/ FT}  & 46.86 &	43.88 &	47.48 &	45.08 &	47.97 &	45.79 &	48.50 &	46.44 &	49.46 &	47.77 &	50.06 &	48.04 &	\underline{49.89} &	\underline{48.47} \\

    & {\bf TFS w/ CT}  & 47.46 &	45.43 &	48.63 &	46.95 &	49.62 &	48.29 &	49.89 &	48.54 &	50.24 &	48.84 &	50.17 &	48.79 & \textbf{50.84} &	\textbf{49.77} \\
    
    \midrule 

    & Random  & 73.98 & 71.69 & 75.91 & 74.21 & 76.62 & 74.93 & 78.11 & 76.84 & 78.97 & 77.94 & 79.33 & 78.49 & 79.95 & 79.09  \\
    & Entropy~\cite{holub2008entropy} & 73.98 & 71.69 & 77.50 & 75.96 & 78.90 & 77.65 & 80.33 & 79.43 & 81.46 & 80.73 & 82.20 & 81.45 & 82.57 & 81.91  \\
    & CoreSet~\cite{sener2018coreset} & 73.98 & 71.69 & 75.81 & 73.95 & 77.76 & 76.24 & 79.16 & 77.94 & 79.80 & 78.58 & 80.74 & 79.77 & 80.99 & 80.16  \\
    & BADGE~\cite{Ash2020Badge} & 73.98 & 71.69 & 77.35 & 75.82 & 78.94 & 77.79 & 80.16 & 79.27 & 81.23 & 80.56 & 82.05 & 81.39 & 82.53 & 81.94  \\
    & PCB + BADGE~\cite{bang2024active} & 73.98 & 71.69 & 76.59 & 74.89 & 78.51 & 77.27 & 80.25 & 79.30 & 81.27 & 80.53 & 81.69 & 81.02 & 82.68 & 82.20  \\
    \textbf{Stanford Cars} & ALFA-Mix~\cite{alfamix}& 73.98 & 71.69 & 76.06 & 74.26 & 77.08 & 75.66 & 78.03 & 76.74 & 78.82 & 77.86 & 79.29 & 78.37 & 79.77 & 78.81  \\
    & LoGo~\cite{kim2023re}  & 73.98 & 71.69 & 76.86 & 75.31 & 78.48 & 77.26 & 80.27 & 79.32 & 81.25 & 80.41 & 81.78 & 81.15 & 82.51 & 81.99  \\
    
    \cmidrule(lr){2-16}

    & {\bf TFS w/ PT} & 73.98 &	71.69 &	77.24 &	75.82 &	78.76 &	77.77 &	79.89 &	79.18 &	81.16 &	80.43 &	81.76 &	81.03 &	82.52 &	82.01  \\

    & {\bf TFS w/ LP}  & 76.11 &	74.20 &	77.98 &	76.59 &	79.56 &	78.51 &	79.59 &	78.51 &	80.05 &	78.98 &	80.65 &	79.65 & 80.96 &	80.00  \\

    & {\bf TFS w/ FT}  & 77.45 &	75.09 &	80.29 &	78.88 &	81.91 &	80.84 &	82.49 &	81.49 &	83.48 &	82.65 &	84.05 &	83.29 &	\underline{84.38} &	\underline{83.61}  \\

    & {\bf TFS w/ CT}  & 83.01 &	82.54 &	83.69 &	83.23 &	84.04 &	83.65 &	84.57 &	84.24 &	84.87 &	84.52 &	85.06 &	84.76 &	\textbf{85.75} &	\textbf{85.47}  \\

    \midrule 
        
    & Random  & 66.08 & 65.08 & 66.25 & 65.19 & 66.86 & 65.91 & 66.66 & 65.78 & 67.45 & 66.55 & 67.62 & 66.84 & 67.58 & 66.88 \\

    & Entropy~\cite{holub2008entropy}  & 66.08 & 65.08 & 66.04 & 65.03 & 67.26 & 66.40 & 67.36 & 66.51 & 67.69 & 66.88 & 67.82 & 67.00 & 68.06 & 67.25 \\
    & CoreSet~\cite{sener2018coreset}  & 66.08 & 65.08 & 65.69 & 64.66 & 66.47 & 65.50 & 66.67 & 65.79 & 66.88 & 66.08 & 66.93 & 66.00 & 67.32 & 66.50 \\
    & BADGE~\cite{Ash2020Badge}  & 66.08 & 65.08 & 66.19 & 65.15 & 67.20 & 66.28 & 67.19 & 66.28 & 67.57 & 66.76 & 68.08 & 67.37 & 68.24 & 67.48 \\
    & PCB + BADGE~\cite{bang2024active} & 66.08 & 65.08 & 65.99 & 64.93 & 66.91 & 65.95 &	66.99 &	66.10 &	67.29 &	66.38 &	67.58 &	66.68 &	68.08 &	67.33 \\
    \textbf{Food101} & ALFA-Mix~\cite{alfamix}  & 66.08 & 65.08  & 66.07 & 65.18 & 66.57 &	65.66 &	66.94 &	66.01 &	66.98 &	66.23 &	67.86 &	67.08 &	67.82 &	67.18 \\
    & LoGo~\cite{kim2023re}  & 66.08 & 65.08 & 66.91 &	65.97 &	67.06 &	66.15 &	67.23 &	66.42 &	67.89 &	67.29 &	68.08 &	67.48 &	68.32 &	67.80 \\
    \cmidrule(lr){2-16}
    & {\bf TFS w/ PT} & 66.08 &	65.08 &	65.72 &	64.63 &	66.65 &	65.71 &	66.41 &	65.45 &	66.97 &	66.04 &	67.30 &	66.29 &	68.34 &	67.54 \\

    & {\bf TFS w/ LP}  & 75.54 &	75.30 &	75.63 &	75.35 &	75.71 &	75.43 &	75.90 &	75.63 &	75.80 &	75.50 &	75.86 &	75.62 &	\textbf{75.77} &	\textbf{75.54} \\

    & {\bf TFS w/ FT}  & 72.65 &	71.99 &	72.87 &	72.12 &	72.93 &	72.28 &	72.95 &	72.24 &	73.03 &	72.37 &	73.16 &	72.48 &	\underline{73.09} &	\underline{72.41} \\
    
    & {\bf TFS w/ CT}  & 72.35 &	71.88 &	72.59 &	72.06 &	72.93 &	72.45 &	72.30 &	71.69 &	72.22 &	71.65 &	72.68 &	72.08 &	72.86 &	72.33 \\
    
    \midrule 

    & Random  & 78.79 &	78.16 &	79.65 &	79.18 &	79.34 &	78.86 &	80.13 &	79.75 &	79.98 &	79.41 &	79.48 &	79.00 &	79.92 &	79.51   \\
    & Entropy~\cite{holub2008entropy} & 78.79 &	78.16 &	79.86 &	79.47 &	79.96 &	79.56 &	80.14 &	79.81 &	80.55 &	80.27 &	81.19 &	80.87 &	80.96 &	80.67   \\
    & CoreSet~\cite{sener2018coreset} & 78.79 &	78.16 &	79.52 &	78.93 &	79.89 &	79.33 &	79.83 &	79.30 &	79.50 &	78.88 &	79.79 &	79.22 &	79.71 &	79.13  \\
    & BADGE~\cite{Ash2020Badge} & 78.79 &	78.16 &	80.10 &	79.67 &	80.20 &	79.89 &	79.58 &	79.13 &	80.70 &	80.35 &	81.08 &	80.74 &	80.90 &	80.60  \\
    & PCB + BADGE~\cite{bang2024active} & 78.79 &	78.16 &	80.04 &	79.65 &	80.35 &	79.95 &	80.40 &	80.09 &	80.49 &	80.11 &	80.88 &	80.51 &	80.80 &	80.45  \\
    \textbf{Oxford Pets} & ALFA-Mix~\cite{alfamix} & 78.79 &	78.16 &	79.79 &	79.29 &	79.62 &	79.25 &	79.83 &	79.44 &	80.19 &	79.89 &	80.15 &	79.90 &	80.76 &	80.48 \\
    & LoGo~\cite{kim2023re}  & 78.79 &	78.16 &	78.93 &	78.29 &	79.59 &	79.12 &	80.74 &	80.36 &	80.97 &	80.70 &	80.33 &	79.93 &	81.17 &	80.92  \\
    \cmidrule(lr){2-16}

    & {\bf TFS w/ PT}  & 78.79 &	78.16 &	79.69 &	79.22 &	80.11 &	79.63 &	80.19 &	79.77 &	80.78 &	80.41 &	80.51 &	80.11 &	81.12 &	80.82  \\

    & {\bf TFS w/ LP} & 87.57 &	87.49 &	87.76 &	87.54 &	87.57 &	87.44 &	87.65 &	87.58 &	87.59 &	87.43 &	87.22 &	87.14 &	\textbf{87.64} &	\textbf{87.57}  \\

    & {\bf TFS w/ FT} & 82.80 &	81.71 &	84.21 &	83.81 &	84.13 &	83.48 &	83.93 &	83.36 &	84.67 &	84.15 &	84.55 &	84.26 &	85.04 &	84.74  \\

    & {\bf TFS w/ CT} & 85.14 & 84.52 &	82.85 &	82.30 &	83.61 &	83.12 &	83.96 &	83.50 &	84.73 &	84.36 &	85.08 &	84.74 &	\underline{85.70} &	\underline{85.44}  \\

\bottomrule

\end{tabular}
}
\label{tab:five_datasets_rda_suppl} 
\end{table*}%
}

{
\setlength{\tabcolsep}{0.2em}
\begin{table*}[t]
\centering
\caption{\small
\textbf{Comparative Analysis of AL methods without RDA.} We systematically evaluate AL methods across five benchmark datasets across five benchmark datasets without RDA. The results below are averaged over three random runs. The complete results for each seed showed in~\Cref{tab:five_datasets_without_rda_suppl_666,tab:five_datasets_without_rda_suppl_777,tab:five_datasets_without_rda_suppl_888}. 
}
\vspace{-2mm}
\scalebox{0.62}
{
\begin{tabular}{llccccccccccccccc}
\toprule
    &
    & \multicolumn{2}{c}{Round 0} 
    & \multicolumn{2}{c}{Round 1} 
    & \multicolumn{2}{c}{Round 2} 
    & \multicolumn{2}{c}{Round 3}  
    & \multicolumn{2}{c}{Round 4} 
    & \multicolumn{2}{c}{Round 5}
    & \multicolumn{2}{c}{Round 6}
    \\

    \cmidrule(lr){3-4} 
    \cmidrule(lr){5-6}
    \cmidrule(lr){7-8} 
    \cmidrule(lr){9-10} 
    \cmidrule(lr){11-12} 
    \cmidrule(lr){13-14} 
    \cmidrule(lr){15-16}

    \textbf{Dataset} &  \textbf{Method} & Acc & Macro F1  & Acc & Macro F1  & Acc & Macro F1  & Acc & Macro F1 & Acc & Macro F1 & Acc & Macro F1 & Acc & Macro F1 \\
    \cmidrule(lr){1-1}
    \cmidrule(lr){2-2}
    \cmidrule(lr){3-4} 
    \cmidrule(lr){5-6}
    \cmidrule(lr){7-8} 
    \cmidrule(lr){9-10} 
    \cmidrule(lr){11-12} 
    \cmidrule(lr){13-14} 
    \cmidrule(lr){15-16} 

      & Random & 11.72 &	10.26 &	18.29 &	16.33 &	22.64 &	20.46 &	26.04 &	24.45 &	28.82 &	27.21 &	31.69 &	30.21 &	33.34 &	32.28 \\
    %\midrule
      & Entropy~\cite{holub2008entropy} & 11.72 &	10.26 &	17.58 &	15.53 &	21.80 &	19.86 &	25.03 &	23.59 &	28.63 &	27.51 &	30.89 &	29.80 &	32.98 &	32.54 \\
      & CoreSet~\cite{sener2018coreset} & 11.72 &	10.26 &	16.77 &	14.83 &	20.84 &	18.57 &	23.90 &	21.87 &	28.07 &	26.09 &	31.17 &	29.69 &	32.53 &	31.06 \\
      \textbf{Semi-Aves}    & BADGE~\cite{Ash2020Badge} & 11.72 &	10.26 &	17.67 &	15.28 &	22.23 &	20.17 &	26.11 &	24.33 &	28.33 &	26.84 &	31.96 &	31.08 &	33.68 &	32.88 \\
      & PCB + BADGE~\cite{bang2024active} & 11.72 &	10.26 &	17.89 &	16.06 &	22.86 &	20.90 &	26.68 &	25.50 &	29.35 &	27.95 &	32.00 &	31.02 &	34.29 &	33.48 \\
      & ALFA-Mix~\cite{alfamix} & 11.72 & 10.26 &	19.20 &	16.84 &	23.95 &	21.46 &	26.81 &	24.97 &	29.87 &	28.47 &	32.05 &	31.01 &	33.62 &	32.54 \\
\bottomrule
\end{tabular}
}
\label{tab:five_datasets_without_rda_suppl} 
\end{table*}%
}

{
\setlength{\tabcolsep}{0.3em}
\begin{table*}[t]
\centering
\caption{\small
\textbf{Comparative Analysis of TFS and AL methods with OpenCLIP (ViT-B/32).} We present a detailed comparison of TFS against established AL methods across five benchmarks (seed = 666). 
}
\vspace{-2mm}
\scalebox{0.61}
{
\begin{tabular}{llccccccccccccccc}
\toprule
    &
    & \multicolumn{2}{c}{Round 0} 
    & \multicolumn{2}{c}{Round 1} 
    & \multicolumn{2}{c}{Round 2} 
    & \multicolumn{2}{c}{Round 3}  
    & \multicolumn{2}{c}{Round 4} 
    & \multicolumn{2}{c}{Round 5}
    & \multicolumn{2}{c}{Round 6}
    \\
    \cmidrule(lr){3-4} 
    \cmidrule(lr){5-6}
    \cmidrule(lr){7-8} 
    \cmidrule(lr){9-10} 
    \cmidrule(lr){11-12} 
    \cmidrule(lr){13-14} 
    \cmidrule(lr){15-16} 
    
    \textbf{Dataset} & \textbf{Method} & Acc & Macro F1  & Acc & Macro F1  & Acc & Macro F1  & Acc & Macro F1 & Acc & Macro F1 & Acc & Macro F1 & Acc & Macro F1 \\
    \cmidrule(lr){1-1}
    \cmidrule(lr){2-2}
    \cmidrule(lr){3-4} 
    \cmidrule(lr){5-6}
    \cmidrule(lr){7-8} 
    \cmidrule(lr){9-10} 
    \cmidrule(lr){11-12} 
    \cmidrule(lr){13-14} 
    \cmidrule(lr){15-16} 

    & Random  & 47.31 & 46.22 & 48.06 & 46.95 & 48.58 & 47.93 & 48.38 & 47.67 & 49.39 & 48.57 & 49.84 & 49.30 & 49.51 & 48.85 \\
    & Entropy~\cite{holub2008entropy}  & 47.31 & 46.22 & 47.70 & 46.50 & 48.18 & 47.26 & 49.39 & 49.00 & 49.71 & 48.59 & 51.23 & 50.64 & 51.04 & 50.26 \\
    & CoreSet~\cite{sener2018coreset}  & 47.31 & 46.22 & 48.18 & 46.79 & 47.85 & 46.48 & 48.40 & 47.38  & 50.15  & 49.02 & 50.24 & 49.47 & 50.35 & 49.97 \\
    & BADGE~\cite{Ash2020Badge}  & 47.31 & 46.22 & 47.88 & 46.82 & 48.66 & 47.67 & 49.45 & 48.86 & 49.73 & 48.71 & 50.15 & 49.58 & 50.44 & 49.70 \\
    & PCB + BADGE~\cite{bang2024active} & 47.31 & 46.22 & 48.60 & 47.36 & 48.41 & 47.18 & 48.81 & 47.92 & 50.54 & 49.72 & 50.98 & 50.44 & 50.14 & 49.76 \\
    \textbf{Semi-Aves} & ALFA-Mix~\cite{alfamix}  & 47.31 & 46.22 & 47.33 & 46.47 & 48.54 & 47.51 & 49.80 & 49.27 & 49.43 & 48.97 & 50.19 & 49.58 & 50.56 & 50.15 \\
\bottomrule

\end{tabular}
}
\label{tab:five_datasets_rda_suppl_seed666} 
\end{table*}%
}

{
\setlength{\tabcolsep}{0.3em}
\begin{table*}[t]
\centering
\caption{\small
\textbf{Comparative Analysis of TFS and AL methods with OpenCLIP (ViT-B/32).} We present a detailed comparison of TFS against established AL methods across five benchmarks (seed = 777). 
}
\vspace{-2mm}
\scalebox{0.61}
{
\begin{tabular}{llccccccccccccccc}
\toprule
    &
    & \multicolumn{2}{c}{Round 0} 
    & \multicolumn{2}{c}{Round 1} 
    & \multicolumn{2}{c}{Round 2} 
    & \multicolumn{2}{c}{Round 3}  
    & \multicolumn{2}{c}{Round 4} 
    & \multicolumn{2}{c}{Round 5}
    & \multicolumn{2}{c}{Round 6}
    \\
    \cmidrule(lr){3-4} 
    \cmidrule(lr){5-6}
    \cmidrule(lr){7-8} 
    \cmidrule(lr){9-10} 
    \cmidrule(lr){11-12} 
    \cmidrule(lr){13-14} 
    \cmidrule(lr){15-16} 
    
    \textbf{Dataset} & \textbf{Method} & Acc & Macro F1  & Acc & Macro F1  & Acc & Macro F1  & Acc & Macro F1 & Acc & Macro F1 & Acc & Macro F1 & Acc & Macro F1 \\
    \cmidrule(lr){1-1}
    \cmidrule(lr){2-2}
    \cmidrule(lr){3-4} 
    \cmidrule(lr){5-6}
    \cmidrule(lr){7-8} 
    \cmidrule(lr){9-10} 
    \cmidrule(lr){11-12} 
    \cmidrule(lr){13-14} 
    \cmidrule(lr){15-16} 

    & Random  & 47.25	&	45.88	&	48.35	&	47.12	&	48.41	&	47.28	&	49.04	&	48.19	&	49.09	&	48.29	&	49.36	&	48.72	&	50.64	&	50.08 \\
    & Entropy~\cite{holub2008entropy}  & 47.25	&	45.88	&	48.19	&	47.21	&	48.46	&	47.53	&	50.10	&	49.12	&	49.36	&	48.56	&	50.21	&	49.65	&	51.01	&	50.42 \\
    & CoreSet~\cite{sener2018coreset}  & 47.25	&	45.88	&	48.33	&	47.28	&	47.78	&	46.36	&	49.05	&	47.98	&	49.51	&	48.70	&	49.54	&	48.57	&	49.99	&	49.05 \\
    & BADGE~\cite{Ash2020Badge}  & 47.25	&	45.88	&	48.61	&	47.64	&	48.78	&	47.86	&	49.90	&	49.03	&	49.90	&	49.07	&	50.41	&	49.69	&	50.75	&	50.11 \\
    & PCB + BADGE~\cite{bang2024active} & 47.25	&	45.88	&	48.53	&	47.56	&	48.93	&	48.03	&	48.70	&	47.96	&	48.99	&	48.02	&	49.73	&	49.00	&	51.15	&	50.30 \\
    \textbf{Semi-Aves} & ALFA-Mix~\cite{alfamix}  & 47.25	&	45.88	&	48.65	&	47.79	&	48.43	&	47.41	&	48.79	&	47.99	&	48.99	&	48.29	&	49.88	&	49.27	&	50.29	&	49.81 \\
\bottomrule

\end{tabular}
}
\label{tab:five_datasets_rda_suppl_seed777} 
\end{table*}%
}

{
\setlength{\tabcolsep}{0.3em}
\begin{table*}[t]
\centering
\caption{\small
\textbf{Comparative Analysis of TFS and AL methods with OpenCLIP (ViT-B/32).} We present a detailed comparison of TFS against established AL methods across five benchmarks (seed = 888). 
}
\vspace{-2mm}
\scalebox{0.61}
{
\begin{tabular}{llccccccccccccccc}
\toprule
    &
    & \multicolumn{2}{c}{Round 0} 
    & \multicolumn{2}{c}{Round 1} 
    & \multicolumn{2}{c}{Round 2} 
    & \multicolumn{2}{c}{Round 3}  
    & \multicolumn{2}{c}{Round 4} 
    & \multicolumn{2}{c}{Round 5}
    & \multicolumn{2}{c}{Round 6}
    \\
    \cmidrule(lr){3-4} 
    \cmidrule(lr){5-6}
    \cmidrule(lr){7-8} 
    \cmidrule(lr){9-10} 
    \cmidrule(lr){11-12} 
    \cmidrule(lr){13-14} 
    \cmidrule(lr){15-16} 
    
    \textbf{Dataset} & \textbf{Method} & Acc & Macro F1  & Acc & Macro F1  & Acc & Macro F1  & Acc & Macro F1 & Acc & Macro F1 & Acc & Macro F1 & Acc & Macro F1 \\
    \cmidrule(lr){1-1}
    \cmidrule(lr){2-2}
    \cmidrule(lr){3-4} 
    \cmidrule(lr){5-6}
    \cmidrule(lr){7-8} 
    \cmidrule(lr){9-10} 
    \cmidrule(lr){11-12} 
    \cmidrule(lr){13-14} 
    \cmidrule(lr){15-16} 

    & Random  & 46.98	&	45.99	&	48.33	&	47.11	&	48.74	&	47.71	&	48.63	&	47.93	&	49.60	&	49.31	&	49.98	&	49.31	&	50.11	&	49.79 \\
    & Entropy~\cite{holub2008entropy}  & 46.98	&	45.99	&	48.36	&	47.42	&	49.33	&	48.44	&	48.96	&	48.47	&	50.25	&	49.30	&	50.84	&	50.53	&	51.01	&	50.82 \\
    & CoreSet~\cite{sener2018coreset}  & 46.98	&	45.99	&	48.08	&	46.78	&	48.89	&	47.74	&	50.01	&	49.07	&	49.50	&	48.66	&	50.50	&	49.70	&	50.55	&	49.70 \\
    & BADGE~\cite{Ash2020Badge}  & 46.98	&	45.99	&	48.25	&	47.26	&	49.44	&	48.49	&	49.30	&	48.79	&	50.45	&	49.64	&	50.34	&	49.77	&	51.01	&	50.73 \\
    & PCB + BADGE~\cite{bang2024active} & 46.98	&	45.99	&	48.53	&	47.59	&	49.13	&	48.34	&	49.54	&	48.73	&	50.38	&	49.77	&	50.74	&	50.57	&	50.94	&	50.58 \\
    \textbf{Semi-Aves} & ALFA-Mix~\cite{alfamix}  & 46.98	&	45.99	&	48.35	&	47.39	&	49.14	&	48.15	&	49.80	&	49.28	&	49.61	&	49.10	&	50.40	&	49.84	&	50.23	&	49.78 \\
\bottomrule

\end{tabular}
}
\label{tab:five_datasets_rda_suppl_seed888} 
\end{table*}%
}

{
\setlength{\tabcolsep}{0.2em}
\begin{table*}[t]
\centering
\caption{\small
\textbf{Comparative Analysis of AL methods without RDA.} We systematically evaluate AL methods across five benchmark datasets across five benchmark datasets without RDA (seed = 666). 
}
\vspace{-2mm}
\scalebox{0.62}
{
\begin{tabular}{llccccccccccccccc}
\toprule
    &
    & \multicolumn{2}{c}{Round 0} 
    & \multicolumn{2}{c}{Round 1} 
    & \multicolumn{2}{c}{Round 2} 
    & \multicolumn{2}{c}{Round 3}  
    & \multicolumn{2}{c}{Round 4} 
    & \multicolumn{2}{c}{Round 5}
    & \multicolumn{2}{c}{Round 6}
    \\

    \cmidrule(lr){3-4} 
    \cmidrule(lr){5-6}
    \cmidrule(lr){7-8} 
    \cmidrule(lr){9-10} 
    \cmidrule(lr){11-12} 
    \cmidrule(lr){13-14} 
    \cmidrule(lr){15-16}

    \textbf{Dataset} &  \textbf{Method} & Acc & Macro F1  & Acc & Macro F1  & Acc & Macro F1  & Acc & Macro F1 & Acc & Macro F1 & Acc & Macro F1 & Acc & Macro F1 \\
    \cmidrule(lr){1-1}
    \cmidrule(lr){2-2}
    \cmidrule(lr){3-4} 
    \cmidrule(lr){5-6}
    \cmidrule(lr){7-8} 
    \cmidrule(lr){9-10} 
    \cmidrule(lr){11-12} 
    \cmidrule(lr){13-14} 
    \cmidrule(lr){15-16} 

      & Random & 11.58	&	10.39	&	18.29	&	16.45	&	22.28	&	20.56	&	26.36	&	24.86	&	29.49	&	28.29	&	32.46	&	31.20	&	33.63	&	32.65 \\
    %\midrule
      & Entropy~\cite{holub2008entropy} & 11.58	&	10.39	&	17.70	&	15.44	&	21.85	&	19.81	&	24.85	&	23.52	&	28.70	&	27.60	&	30.13	&	28.92	&	32.83	&	32.44 \\
      & CoreSet~\cite{sener2018coreset} & 11.58	&	10.39	&	16.93	&	14.81	&	20.78	&	18.33	&	24.24	&	22.39	&	28.09	&	26.38	&	30.89	&	29.31	&	32.13	&	30.56 \\
      \textbf{Semi-Aves}    & BADGE~\cite{Ash2020Badge} & 11.58	&	10.39	&	17.44	&	15.18	&	22.25	&	20.28	&	26.03	&	24.14	&	27.94	&	26.62	&	32.13	&	30.82	&	32.90	&	32.01 \\
      & PCB + BADGE~\cite{bang2024active} & 11.58	&	10.39	&	18.54	&	16.41	&	24.05	&	21.82	&	27.98	&	26.76	&	29.95	&	28.66	&	33.09	&	32.47	&	34.86	&	34.32 \\
      & ALFA-Mix~\cite{alfamix} & 11.58	&	10.39	&	19.88	&	17.59	&	23.80	&	21.46	&	26.10	&	24.18	&	29.36	&	28.05	&	31.89	&	30.77	&	33.46	&	32.52 \\
\bottomrule
\end{tabular}
}
\label{tab:five_datasets_without_rda_suppl_666} 
\end{table*}%
}

{
\setlength{\tabcolsep}{0.2em}
\begin{table*}[t]
\centering
\caption{\small
\textbf{Comparative Analysis of AL methods without RDA.} We systematically evaluate AL methods across five benchmark datasets across five benchmark datasets without RDA (seed = 777). 
}
\vspace{-2mm}
\scalebox{0.62}
{
\begin{tabular}{llccccccccccccccc}
\toprule
    &
    & \multicolumn{2}{c}{Round 0} 
    & \multicolumn{2}{c}{Round 1} 
    & \multicolumn{2}{c}{Round 2} 
    & \multicolumn{2}{c}{Round 3}  
    & \multicolumn{2}{c}{Round 4} 
    & \multicolumn{2}{c}{Round 5}
    & \multicolumn{2}{c}{Round 6}
    \\

    \cmidrule(lr){3-4} 
    \cmidrule(lr){5-6}
    \cmidrule(lr){7-8} 
    \cmidrule(lr){9-10} 
    \cmidrule(lr){11-12} 
    \cmidrule(lr){13-14} 
    \cmidrule(lr){15-16}

    \textbf{Dataset} &  \textbf{Method} & Acc & Macro F1  & Acc & Macro F1  & Acc & Macro F1  & Acc & Macro F1 & Acc & Macro F1 & Acc & Macro F1 & Acc & Macro F1 \\
    \cmidrule(lr){1-1}
    \cmidrule(lr){2-2}
    \cmidrule(lr){3-4} 
    \cmidrule(lr){5-6}
    \cmidrule(lr){7-8} 
    \cmidrule(lr){9-10} 
    \cmidrule(lr){11-12} 
    \cmidrule(lr){13-14} 
    \cmidrule(lr){15-16} 

      & Random & 11.79	&	10.20	&	18.33	&	16.14	&	23.08	&	20.80	&	26.36	&	24.62	&	28.49	&	26.59	&	31.46	&	29.86	&	33.03	&	31.87 \\
    %\midrule
      & Entropy~\cite{holub2008entropy} & 11.79	&	10.20	&	16.91	&	15.16	&	21.60	&	19.36	&	24.75	&	23.17	&	28.01	&	26.60	&	31.13	&	29.80	&	33.06	&	32.72 \\
      & CoreSet~\cite{sener2018coreset} & 11.79	&	10.20	&	16.81	&	15.16	&	20.98	&	18.42	&	23.63	&	21.29	&	28.01	&	25.52	&	30.96	&	29.55	&	31.76	&	30.57 \\
      \textbf{Semi-Aves}    & BADGE~\cite{Ash2020Badge} & 11.79	&	10.20	&	16.71	&	14.93	&	21.69	&	19.25	&	25.34	&	23.64	&	27.44	&	25.64	&	31.15	&	30.38	&	32.99	&	32.16 \\
      & PCB + BADGE~\cite{bang2024active} & 11.79	&	10.20	&	16.55	&	14.95	&	21.88	&	20.10	&	25.20	&	24.06	&	28.75	&	27.23	&	30.76	&	29.41	&	33.55	&	32.40 \\
      & ALFA-Mix~\cite{alfamix} & 11.79	&	10.20	&	18.33	&	16.06	&	24.30	&	21.50	&	27.28	&	25.35	&	29.60	&	28.11	&	32.69	&	31.79	&	33.05	&	31.98 \\
\bottomrule
\end{tabular}
}
\label{tab:five_datasets_without_rda_suppl_777} 
\end{table*}%
}

{
\setlength{\tabcolsep}{0.2em}
\begin{table*}[t]
\centering
\caption{\small
\textbf{Comparative Analysis of AL methods without RDA.} We systematically evaluate AL methods across five benchmark datasets across five benchmark datasets without RDA (seed = 888). 
}
\vspace{-2mm}
\scalebox{0.62}
{
\begin{tabular}{llccccccccccccccc}
\toprule
    &
    & \multicolumn{2}{c}{Round 0} 
    & \multicolumn{2}{c}{Round 1} 
    & \multicolumn{2}{c}{Round 2} 
    & \multicolumn{2}{c}{Round 3}  
    & \multicolumn{2}{c}{Round 4} 
    & \multicolumn{2}{c}{Round 5}
    & \multicolumn{2}{c}{Round 6}
    \\

    \cmidrule(lr){3-4} 
    \cmidrule(lr){5-6}
    \cmidrule(lr){7-8} 
    \cmidrule(lr){9-10} 
    \cmidrule(lr){11-12} 
    \cmidrule(lr){13-14} 
    \cmidrule(lr){15-16}

    \textbf{Dataset} &  \textbf{Method} & Acc & Macro F1  & Acc & Macro F1  & Acc & Macro F1  & Acc & Macro F1 & Acc & Macro F1 & Acc & Macro F1 & Acc & Macro F1 \\
    \cmidrule(lr){1-1}
    \cmidrule(lr){2-2}
    \cmidrule(lr){3-4} 
    \cmidrule(lr){5-6}
    \cmidrule(lr){7-8} 
    \cmidrule(lr){9-10} 
    \cmidrule(lr){11-12} 
    \cmidrule(lr){13-14} 
    \cmidrule(lr){15-16} 

      & Random & 11.79	&	10.20	&	18.24	&	16.40	&	22.55	&	20.01	&	25.40	&	23.87	&	28.48	&	26.76	&	31.15	&	29.56	&	33.38	&	32.32 \\
    %\midrule
      & Entropy~\cite{holub2008entropy} & 11.79	&	10.20	&	18.13	&	15.98	&	21.95	&	20.42	&	25.48	&	24.06	&	29.19	&	28.33	&	31.40	&	30.67	&	33.05	&	32.46  \\
      & CoreSet~\cite{sener2018coreset} & 11.79	&	10.20	&	16.59	&	14.51	&	20.75	&	18.97	&	23.84	&	21.94	&	28.13	&	26.38	&	31.68	&	30.22	&	33.70	&	32.04  \\
      \textbf{Semi-Aves}    & BADGE~\cite{Ash2020Badge} & 11.79	&	10.20	&	18.86	&	15.73	&	22.74	&	20.99	&	26.95	&	25.20	&	29.60	&	28.27	&	32.60	&	32.02	&	35.14	&	34.48 \\
      & PCB + BADGE~\cite{bang2024active} & 11.79	&	10.20	&	18.59	&	16.82	&	22.64	&	20.78	&	26.86	&	25.69	&	29.36	&	27.96	&	32.15	&	31.19	&	34.45	&	33.72 \\
      & ALFA-Mix~\cite{alfamix} & 11.79	&	10.20	&	19.39	&	16.89	&	23.75	&	21.44	&	27.05	&	25.37	&	30.65	&	29.25	&	31.58	&	30.47	&	34.35	&	33.11 \\
\bottomrule
\end{tabular}
}
\label{tab:five_datasets_without_rda_suppl_888} 
\end{table*}%
}

\end{document}